\documentclass[10pt,twocolumn,letterpaper]{article}
\usepackage[table,xcdraw,dvipsnames]{xcolor}

\DeclareRobustCommand*{\ora}{\overrightarrow}
\DeclareRobustCommand*{\ola}{\overleftarrow}

\definecolor{col1}{RGB}{232, 161, 148}
\definecolor{col2}{RGB}{148, 187, 232}

\usepackage{cvpr}
\usepackage{times}
\usepackage{epsfig}
\usepackage{graphicx}
\usepackage{amsmath}
\usepackage{amssymb}
\usepackage[percent]{overpic}
\usepackage{pdfpages}
\usepackage{multido}

\usepackage[pagebackref=true,breaklinks=true,letterpaper=true,colorlinks,bookmarks=false]{hyperref}

\cvprfinalcopy 

\def\cvprPaperID{2583} 

\ifcvprfinal\fi
\begin{document}

\title{On the uncertainty of self-supervised monocular depth estimation}

\author{Matteo Poggi \hspace*{1cm} Filippo Aleotti \hspace*{1cm} Fabio Tosi \hspace*{1cm} Stefano Mattoccia\\
Department of Computer Science and Engineering (DISI)\\
University of Bologna, Italy\\
{\tt\small \{m.poggi, filippo.aleotti2, fabio.tosi5, stefano.mattoccia \}@unibo.it}
}

\maketitle
\thispagestyle{empty}

\begin{abstract}
Self-supervised paradigms for monocular depth estimation are very appealing since they do not require ground truth annotations at all. Despite the astonishing results yielded by such methodologies, learning to reason about the uncertainty of the estimated depth maps is of paramount importance for practical applications, yet uncharted in the literature. Purposely, we explore for the first time how to estimate the uncertainty for this task and how this affects depth accuracy, proposing a novel peculiar technique specifically designed for self-supervised approaches. On the standard KITTI dataset, we exhaustively assess the performance of each method with different self-supervised paradigms. Such evaluation highlights that our proposal i) always improves depth accuracy significantly and ii) yields state-of-the-art results concerning uncertainty estimation when training on sequences and competitive results uniquely deploying stereo pairs. 

\end{abstract}

\section{Introduction}

Depth estimation is often pivotal to a variety of high-level tasks in computer vision, such as autonomous driving, augmented reality, and more. Although active sensors such as LiDAR  are deployed for some of the applications mentioned above, estimating depth from standard cameras is generally preferable due to several advantages. Among them: the much lower cost of standard imaging devices, their higher resolution and frame rate allow for more scalable and compelling solutions.
\begin{figure}
    \centering
    \begin{tabular}{c}
        \includegraphics[width=0.4\textwidth]{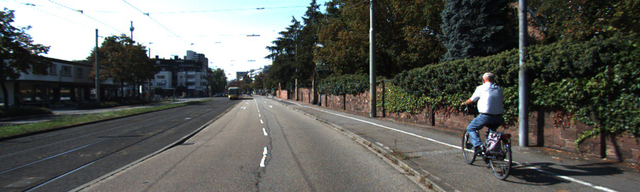} \\
        \includegraphics[width=0.4\textwidth]{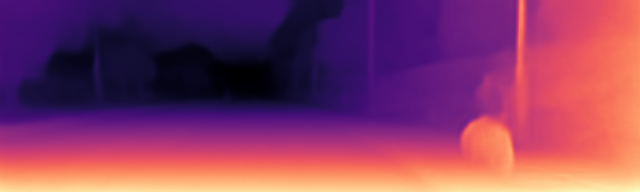} \\
        \begin{overpic}[width=0.4\textwidth]{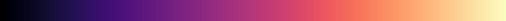}
        \put (1,1) {$\displaystyle\fontsize{8}{8}\textcolor{green}{\text{Far}}$}
        \put (90,1) {$\displaystyle\fontsize{8}{8}\textcolor{red}{\text{Close}}$}
        \end{overpic} \\
        \includegraphics[width=0.4\textwidth]{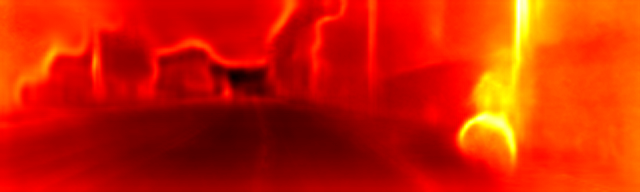} \\
        \begin{overpic}[width=0.4\textwidth]{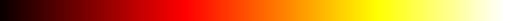}
        \put (1,1) {$\displaystyle\fontsize{8}{8}\textcolor{green}{\text{Low}}$}
        \put (90,1) {$\displaystyle\fontsize{8}{8}\textcolor{red}{\text{High}}$}
        \end{overpic} \\
    \end{tabular}
    \caption{\textbf{How much can we trust self-supervised monocular depth estimation?} From a single input image (top) we estimate depth (middle) and uncertainty (bottom) maps. Best with colors.}
    \label{fig:abstract}
\end{figure}

In computer vision, depth perception from two \cite{scharstein2002taxonomy} or multiple images \cite{seitz2006comparison} has a long history. Nonetheless, only in the last decade depth estimation from a single image \cite{saxena2008make3d} became an active research topic. On the one hand, this direction is particularly attractive because it overcomes several limitations of the traditional multi-view solutions (\eg, occlusions, overlapping framed area, and more), enabling depth perception with any device equipped with a camera. Unfortunately, it is an extremely challenging task due to the ill-posed nature of the problem. 

Deep learning ignited the spread of depth-from-mono frameworks \cite{eigen2014depth,laina2016deeper,fu2018supervised}, at the cost of requiring a large number of image samples annotated with ground truth depth labels \cite{SilbermanECCV12,uhrig2017sparsity} to achieve satisfying results. However, sourcing annotated depth data is particularly expensive and cumbersome. Indeed, in contrast to many other supervised tasks for which offline handmade annotation is tedious, yet relatively easy, gathering accurate depth labels requires active (and often expensive) sensors and specific calibration, making offline annotation hardly achievable otherwise. Self-supervised \cite{monodepth17, zhou2017unsupervised, mahjourian2018unsupervised, 3net18, pydnet18} or weakly supervised \cite{yang2018deep,Tosi_2019_CVPR,watson2019hints} paradigms, leveraging on image reprojection and noisy labels respectively, 
have removed this issue and yield accuracy close to supervised methods \cite{fu2018supervised}, neglecting at all the deployment of additional depth sensors for labeling purposes. Among self-supervised paradigms, those deploying monocular sequences are more challenging since scale and camera poses are unknown, yet preferred for most practical applications since they allow gathering of training data with the same device used to infer depth.

As for other perception strategies, it is essential to find out failure cases, when occurring, in monocular depth estimation networks. For instance, in an autonomous driving scenario, the erroneous perception of the distance to pedestrians or other vehicles might have dramatic consequences. Moreover, the ill-posed nature of depth-from-mono perception task makes this eventuality much more likely to occur compared to techniques leveraging scene geometry \cite{scharstein2002taxonomy,seitz2006comparison}. In these latter cases, estimating the \emph{uncertainty} (or, complementary, the \emph{confidence}) proved to be effective for depth-from-stereo, by means of both model-based \cite{hu2012quantitative} and learning-based \cite{poggi2017quantitative,kendall2018multi} methods, optical flow \cite{ilg2018uncertainty}, and semantic segmentation \cite{huang2018efficient,kendall2018multi}.
Despite the steady progress in other related fields, uncertainty estimation for self-supervised paradigms remains almost unexplored or, when faced, not quantitatively evaluated \cite{Klodt_2018_ECCV}.

Whereas concurrent works in this field \cite{Godard_2019_ICCV,watson2019hints,Tosi_2019_CVPR} targeted uniquely depth accuracy, we take a breath on this rush and focus for the first time, to the best of our knowledge, on uncertainty estimation for self-supervised monocular depth estimation networks, showing how this practise enables to improve depth accuracy as well. 

Our main contributions can be summarized as follows: 

\begin{itemize}
    \item A comprehensive evaluation of uncertainty estimation approaches tailored for the considered task. 
    
    \item An in-depth investigation of how the self-supervised training paradigm deployed impacts uncertainty and depth estimation. 

    \item A new and peculiar \emph{Self-Teaching} paradigm to model uncertainty, particularly useful when the pose is unknown during the training process, always enabling to improve depth accuracy.

\end{itemize}

Deploying standard metrics in this field, we provide exhaustive experimental results on the KITTI dataset \cite{geiger2013vision}. Figure \ref{fig:abstract} shows the output of a state-of-the-art monocular depth estimator network enriched to model uncertainty. We can notice how our proposal effectively allows to detect wrong predictions (\eg, in the proximity of the person riding the bike).

\section{Related work}

In this section, we review the literature concerning self-supervised monocular depth estimation and techniques to estimate uncertainty in deep neural networks.

\textbf{Self-supervision for mono.}
The advent of deep learning, together with the increasing availability of ground truth depth data, led to the development of frameworks \cite{laina2016deeper,liu2016learning, xu2018supervised,fu2018supervised} achieving unpaired accuracy compared to previous approaches \cite{saxena2009make3d,ladicky2014pulling,fanello2014learning}.
Nonetheless, the effort to collect large amounts of labeled images is high. Thus, to overcome the need for ground truth data, self-supervision in the form of image reconstruction represents a prevalent research topic right now. Frameworks leveraging on this paradigm belong to two (not mutually exclusive) categories, respectively supervised through monocular sequences or stereo pairs. 

The first family of networks jointly learns to estimate the depth and relative pose between two images acquired by a moving camera. Seminal work in this direction is \cite{zhou2017unsupervised}, extended by leveraging on point-cloud alignment \cite{mahjourian2018unsupervised}, differentiable DVO \cite{wang2018unsupervised}, optical flow \cite{yin2018geonet,zou2018df,Chen_ICCV_2019,Ranjan_CVPR_2019}, semantic \cite{Tosi_2020_CVPR} or scale consistency \cite{Bian_NeurIPS_2019}. One of the shortcomings of these approaches is represented by moving objects appearing in the training images, addressed in \cite{Casser_AAAI_2019, Xu_IJCAI_2019} employing instance segmentation and subsequent motion estimation of the segmented dynamic objects.

For the second category, pivotal are the works by Garg \etal \cite{garg2016unsupervised} and Godard \etal \cite{monodepth17}. Other methods improved efficiency \cite{pydnet18,DATE_2019} to enable deployment on embedded devices, or accuracy by simulating a trinocular setup \cite{3net18}, jointly learning for semantic \cite{ramirez2018}, using higher resolution \cite{pillai2019superdepth}, GANs \cite{Aleotti_monogan_2018}, sparse inputs from visual odometry \cite{vomonodepth19} or a teacher-student scheme \cite{Pilzer_2019_CVPR}. 
Finally, approaches leveraging both kind of supervisions have been proposed in  \cite{Zhan_CVPR_2018,Yang_ECCV_Workshops_2018,Luo_EPC++_2018,Godard_2019_ICCV}.

\textbf{Weak-supervision for mono.} A trade-off between self and full supervision is represented by another family of approaches leveraging \emph{weaker} annotations.
In this case, labels can be sourced from synthetic datasets \cite{mayer2016large}, used to train stereo networks for single view stereo \cite{luo2018single} and label distillation \cite{guo2018learning} or in alternative to learn depth estimation and perform domain transfer when dealing with real images \cite{atapour2018real}.

Another source of weak supervision consists of using \emph{noisy} annotations obtained employing the raw output of a LiDAR sensor \cite{Kuznietsov_2017_CVPR} or model-based algorithms. In this latter case, the use of conventional stereo algorithms such as SGM \cite{hirschmuller08} to obtain proxy labels \cite{Tosi_2019_CVPR,watson2019hints}, optionally together with confidence measures \cite{tonioni2019unsupervised}, allowed improving self-supervision from stereo pairs. Other works distilled noisy labels leveraging on structure from motion \cite{Klodt_2018_ECCV} or direct stereo odometry \cite{yang2018deep}.

\begin{figure*}
    \centering
    \includegraphics[width=0.95\textwidth]{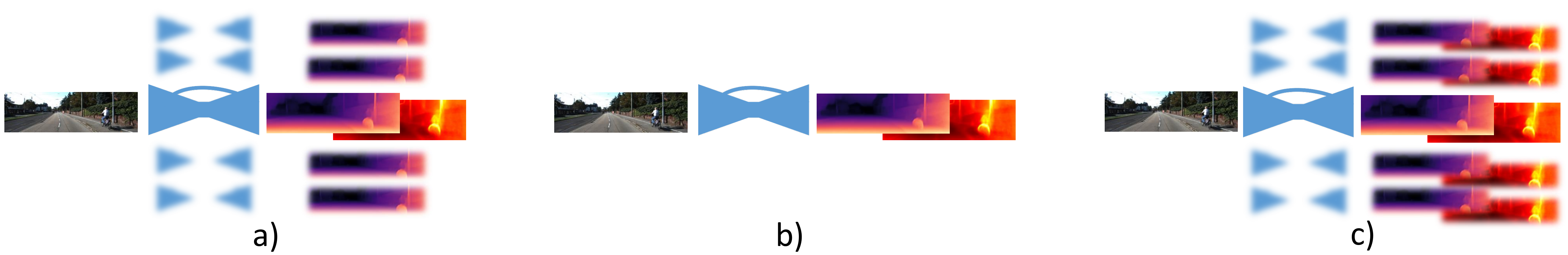}
    \caption{\textbf{Overview of uncertainty estimation implementations.} Respectively a) empirical methods model uncertainty as the variance of predictions from a subset of all the possible instances of the same network, b) predictive are trained to estimate depth and uncertainty as mean and variance of a distribution and c) Bayesian methods are approximated \cite{neal2012bayesian} by sampling multiple predictive models and summing single uncertainties with the variance of the depth predictions.}
    \label{fig:methods}
\end{figure*}

\textbf{Uncertainty estimation.} Estimating the uncertainty (or, complementary, confidence) of cues inferred from images is of paramount importance for their deployment in real computer vision applications. This aspect has been widely explored even before the spread of deep learning, for instance, when dealing with optical flow and stereo matching. 
Concerning optical flow, uncertainty estimation methods belong to two main categories: \emph{model-inherent} and \emph{post-hoc}. The former family \cite{bruhn2006confidence,kybic2011bootstrap,wannenwetsch2017probflow} estimates uncertainty scores based on the internal flow estimation model, \ie, energy minimization models, while the latter \cite{mac2012learning,kondermann2007adaptive,kondermann2008statistical} analyzing already estimated flow fields.
Regarding stereo vision, confidence estimation has been inferred similarly. At first, from features extracted by the internal disparity estimation model, \ie, the cost volume \cite{hu2012quantitative}, then by means of deep learning on already estimated disparity maps \cite{poggi2017quantitative,seki2016patch,poggi2016bmvc,tosi2018beyond,Kim_2019_CVPR}.

Uncertainty estimation has a long history in neural networks as well, starting with Bayesian neural networks \cite{mackay1992practical,chen2014stochastic,welling2011bayesian}. Different models are \textit{sampled} from the distribution of weights to estimate mean and variance of the target distribution in an \emph{empirical} manner. In \cite{graves2011practical,blundell2015weight}, sampling was replaced by variational inference. 
Additional strategies to sample from the distribution of weights are bootstrapped ensembles \cite{lakshminarayanan2017simple} and Monte Carlo Dropout \cite{gal2016dropout}.
A different strategy consists of estimating uncertainty in a \emph{predictive} manner.  Purposely, a neural network is trained to infer the mean and variance of the distribution rather than a single value \cite{nix1994estimating}. This strategy is both effective and cheaper than empirical strategies, since it does not require multiple forward passes and can be adapted to self-supervised approaches as shown in \cite{Klodt_2018_ECCV}. 
Recent works \cite{kendall2017uncertainties,kendall2018multi} combined both in a joint framework.

Finally, Ilg \etal \cite{ilg2018uncertainty} conducted studies about uncertainty modelling for deep optical flow networks. Nonetheless, in addition to the different nature of our task (\ie, the ill-posed monocular depth estimation problem), our work differs for the supervision paradigm, traditional in their case and self-supervised in ours.

\section{Depth-from-mono and uncertainty}

In this section, we introduce how to tackle uncertainty modelling with self-supervised depth estimation frameworks. Given a still image $\mathcal{I}$ any depth-from-mono framework produces an output map $d$ encoding the depth of the observed scene. When full supervision is available, to train such a network we aim at minimizing a loss signal $\mathcal{L}_{fs}$ obtained through a generic function $\mathcal{F}$ of inputs estimated $d$ and ground truth $d^*$ depth maps.

\begin{equation}
    \mathcal{L}_{fs} = \mathcal{F}(d, d^*)
\end{equation}
When traditional supervision is not available, it can be replaced by self-supervision obtained through image reconstruction.
In this case, the ground truth map $d^*$ is replaced by a second image $\mathcal{I}$\textsuperscript{\textdagger}. Then, by knowing camera intrinsics $K$, $K$\textsuperscript{\textdagger} and the relative camera pose $(R|t)$ between the two images, a reconstructed image $\tilde{\mathcal{I}}$ is obtained as a function $\pi$ of intrinsics, pose, image $\mathcal{I}$\textsuperscript{\textdagger} and depth $d$, enabling to compute a loss signal $\mathcal{L}_{ss}$ as a generic $\mathcal{F}$ of inputs $\tilde{\mathcal{I}}$ and $\mathcal{I}$.

\begin{equation}
    \mathcal{L}_{ss} = \mathcal{F}(\tilde{\mathcal{I}}, \mathcal{I}) = \mathcal{F}( \pi(\mathcal{I}\textsuperscript{\textdagger}, K\textsuperscript{\textdagger}
, R|t, K, d), \mathcal{I})
\label{eq:reproj}
\end{equation}
$\mathcal{I}$ and $\mathcal{I}$\textsuperscript{\textdagger} can be acquired either by means of a single moving camera or with a stereo rig. In this latter case, $(R|t)$ is known beforehand thanks to the stereo calibration parameters, while for images acquired by a single camera it is usually learned jointly to depth, both up to a scale factor. A popular choice for $\mathcal{F}$ is a weighted sum between L1 and Structured Similarity Index Measure (SSIM) \cite{SSIM}

\begin{equation}
    \mathcal{F}(\tilde{\mathcal{I}}, \mathcal{I}) = \alpha \cdot \frac{1- \text{SSIM}(\tilde{\mathcal{I}}, \mathcal{I})}{2} + (1-\alpha) \cdot|\tilde{\mathcal{I}}-\mathcal{I}|
\end{equation}
with $\alpha$ commonly set to 0.85 \cite{Godard_2019_ICCV}. In case of $K$ frames used for supervision, coming for example by joint monocular and stereo supervision, for each pixel $q$ the minimum among computed losses allows for robust reprojection \cite{Godard_2019_ICCV}

\begin{equation}
    \mathcal{L}_{ss}(q) = \min_{i \in [0..K]} \mathcal{F}(\tilde{\mathcal{I}}_i(q), \mathcal{I}(q))
\end{equation}

Traditional networks are deterministic, producing a single output typically corresponding to the mean value of the distribution of all possible outputs $p(d^*|\mathcal{I}, \mathcal{D}$), $\mathcal{D}$ being a dataset of images and corresponding depth maps. Estimating the variance of such distribution allows for modelling uncertainty on the network outputs, as shown in \cite{ilg2017flownet,kendall2017uncertainties} and depicted in Figure \ref{fig:methods}, a) in empirical way, b) by learning a predictive model or c) combining the two approaches.

First and foremost, we point out that the self-supervision provided to the network is \emph{indirect} with respect to its main task.  
This means that the network estimates are not optimized with respect to the desired statistical distribution, \ie depth $d^*$, but they are an input parameter of a function ($\pi$) optimized over a different statistical model, \ie image $\mathcal{I}$. While this does not represent an issue for empirical methods, predictive methods like negative log-likelihood minimization can be adapted to this paradigm as done by Klodt and Vedaldi \cite{Klodt_2018_ECCV}. Nevertheless, we will show how this solution is sub-optimal when the pose is unknown, \ie when $\pi$ is function of two unknown parameters. 

\begin{figure}
    \centering
    \includegraphics[width=0.95\linewidth]{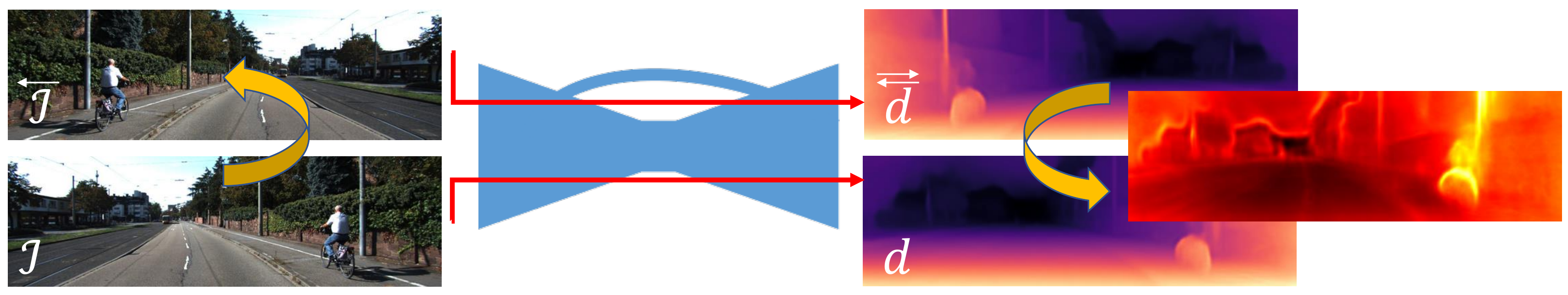}
    \caption{\textbf{Uncertainty by image flipping.} The difference between the depth $d$, inferred from image $\mathcal{I}$, and the depth $\ora{\ola{d}}$, from the flipped image $\ola{\mathcal{I}}$, provides a basic form of uncertainty.}
    \label{fig:post-processing}
\end{figure}

\subsection{Uncertainty by image flipping} 

A simple strategy to estimate uncertainty is inspired by the post-processing (\textit{\textit{Post}}) step proposed by Godard \etal \cite{monodepth17}. Such a refinement consists of estimating two depth maps $d$ and $\ola{d}$ for image $\mathcal{I}$ and its horizontally flipped counterpart $\ola{\mathcal{I}}$. The refined depth map $d^r$ is obtained by averaging $d$ and $\ora{\ola{d}}$, \ie back-flipped $\ola{d}$. We encode the uncertainty for $d^r$ as the difference between the two

\begin{equation}
    u_\text{\textit{\textit{Post}}} = | d - \ora{\ola{d}} |
\end{equation}
\ie, the variance over a small distribution of outputs (\ie, two), as typically done for empirical methods outlined in the next section.
Although this method requires $2\times$ forwards at test time compared to the raw depth-from-mono model, as shown in Figure \ref{fig:post-processing}, it can be applied seamlessly to any pre-trained framework without any modification.

\subsection{Empirical estimation}

This class of methods aims at encoding uncertainty empirically, for instance, by measuring the variance between a set of all the possible network configurations. It allows to explain the model uncertainty, namely \emph{epistemic} \cite{kendall2017uncertainties}. Strategies belonging to this category \cite{ilg2018uncertainty} can be applied to self-supervised frameworks straightforwardly.

\textbf{Dropout Sampling (\textit{Drop}).} Early works estimated uncertainty in neural networks \cite{mackay1992practical} by sampling multiple networks from the distribution of weights of a single architecture. Monte Carlo Dropout \cite{srivastava2014dropout} represents a popular method to sample N independent models without requiring multiple and independent trainings. 
At training time, connections between layers are randomly dropped with a probability $p$ to avoid overfitting. At test time, all connections are kept. By keeping dropout enabled at test time, we can perform multiple forwards sampling a different network every time. Empirical mean $\mu(d)$ and variance $\sigma^2(d)$ are computed, as follows, performing multiple (N) inferences:

\begin{equation}
    \mu(d) = \frac{1}{N} \sum_{i=1}^N d_i
\end{equation}

\begin{equation}
    u_\text{\textit{Drop}} = \sigma^2(d) = \frac{1}{N} \sum_{i=1}^N (d_i - \mu(d))^2
\end{equation}
At test time, using the same number of network parameters, N$\times$ forwards are required.

\textbf{Bootstrapped Ensemble (\textit{Boot}).} A simple, yet effective alternative to weights sampling is represented by training an ensemble of N neural networks \cite{lakshminarayanan2017simple} randomly initializing N instances of the same architecture and training them with bootstrapping, \ie on random subsets of the entire training set. This strategy produces N specialized models. Then, similarly to dropout sampling, we can obtain empirical mean $\mu(d)$ and variance $\sigma^2(d)$ in order to approximate the mean and variance of the distribution of depth values. It requires N$\times$ parameters to be stored, results on N$\times$ independent trainings, and a single forward pass for each stored configuration at test time. 

\textbf{Snapshot Ensemble (\textit{Snap}).} Although the previous method is compelling, obtaining ensembles of neural networks is expensive since it requires carrying out N independent training. An alternative solution \cite{huang2017snapshot} consists of obtaining N snapshots out of a single training by leveraging on cyclic learning rate schedules to obtain $C$ pre-converged models. 
Assuming an initial learning rate $\lambda_0$, we obtain $\lambda_t$ at any training iteration $t$ as a function of the total number of steps $T$ and cycles $C$ as in \cite{huang2017snapshot}

\begin{equation}
    \lambda_t = \frac{\lambda_0}{2} \cdot \left( \cos \left(\frac{\pi \cdot \mod(t-1, \lceil \frac{T}{C} \rceil)}{\lceil \frac{T}{C} \rceil }\right) + 1 \right)
\end{equation}
Similarly to \textit{Boot} and \textit{Drop}, we obtain empirical mean $\mu(d)$ and variance $\sigma^2(d)$ by choosing $N$ out of the $C$ models obtained from a single training procedure.

\subsection{Predictive estimation}

This category aims at encoding uncertainty by learning a predictive model. This means that at test time these methods produce estimates that are function of network parameters and the input image and thus reason about the current observations, modelling \emph{aleatoric heteroscedastic} uncertainty \cite{kendall2017uncertainties}. 
Since often learned from real data distribution, for instance as a function of the distance between the predictions and the ground truth or by maximizing log-likelihood, these approaches need to be rethought to deal with self-supervised paradigms.

\textbf{Learned Reprojection (\textit{Repr}).} To learn a function over the prediction error employing a classifier is a popular technique used for both stereo \cite{poggi2017quantitative,shaked2017improved} and optical flow \cite{mac2012learning}.
However, given the absence of ground truth labels, we cannot apply this approach to self-supervised frameworks seamlessly. Nevertheless, we can drive one output of our network to mimic the behavior of the self-supervised loss function used to train it, thus learning ambiguities affecting the paradigm itself (\eg, occlusions, low texture and more). Indeed, the per-pixel loss signal is supposed to be high when the estimated depth is wrong. Thus, uncertainty $u_\text{\textit{Repr}}$ is trained adding the following term to $\mathcal{L}_{ss}$

\begin{equation}
    \mathcal{L}_\text{\textit{Repr}} = \beta \cdot | u_\text{\textit{Repr}} - \mathcal{F}(\tilde{\mathcal{I}}, \mathcal{I})  |
\end{equation}
Since multiple images $\mathcal{I}$\textsuperscript{\textdagger} may be used for supervision, \ie when combining monocular and stereo, usually for each pixel $q$ the minimum reprojection signal is considered to train the network, thus $u_\text{\textit{Repr}}$ is trained accordingly

\begin{equation}
    \mathcal{L}_\text{\textit{Repr}}(q) = \beta \cdot | u_\text{\textit{Repr}}(q) - \min_{i \in [0..K]}\mathcal{F}(\tilde{\mathcal{I}_i}(q), \mathcal{I}(q)) |
\end{equation}

In our experiments, we set $\beta$ to 0.1 and stop $\mathcal{F}$ gradients inside $\mathcal{L}_\text{\textit{Repr}}$ for numerical stability. A similar technique appeared in \cite{chen2018conf}, although not evaluated quantitatively.

\textbf{Log-Likelihood Maximization (\textit{Log}).} Another popular strategy \cite{nix1994estimating} consists of training the network to infer mean and variance of the distribution $p(d^*|\mathcal{I},\mathcal{D})$ of parameters $\Theta$. The network is trained by log-likelihood maximization (\ie, negative log-likelihood minimization)

\begin{equation}
    \log p(d^*|w) = \frac{1}{N} \sum_q \log p(d^*(q)| \Theta(\mathcal{I},w))
\end{equation}
$w$ being the network weights. As shown in \cite{ilg2018uncertainty}, the predictive distribution can be modelled as Laplacian or Gaussian respectively in case of L1 or L2 loss computation with respect to $d^*$. In the former case, this means minimizing the following loss function

\begin{equation}
    \mathcal{L}_\text{\textit{Log}} = \frac{|\mu(d) - d^*|}{\sigma(d)} + \log \sigma(d)
\end{equation}
with $\mu(d)$ and $\sigma(d)$ outputs of the network encoding mean and variance of the distribution. The additional logarithmic term discourages infinite predictions for any pixel. Regarding numerical stability \cite{kendall2017uncertainties}, the network is trained to estimate the log-variance in order to avoid zero values of the variance.
As shown by Klodt and Vedaldi \cite{Klodt_2018_ECCV}, in absence of ground truth $d^*$ one can model the uncertainty $u_\text{\textit{Log}}$ according to photometric matching

\begin{equation}
    \mathcal{L}_\text{\textit{Log}} = \frac{\min_{i \in [0..K]} \mathcal{F}(\tilde{\mathcal{I}}_i(q), \mathcal{I}(q))}{u_\text{\textit{Log}}} + \log u_\text{\textit{Log}}
\end{equation}
Recall that $\mathcal{F}$ is computed over $\pi$ according to Equation \ref{eq:reproj}. Although for stereo supervision this formulation is equivalent to traditional supervision, \ie $\pi$ is function of a single unknown parameter $d$, in case of monocular supervision this formulation jointly explain uncertainty for depth and pose, both unknown variables in $\pi$. We will show how this approach leads to sub-optimal modelling and how to overcome this limitation with the next approach.

\begin{figure}
    \centering
    \includegraphics[width=0.95\linewidth]{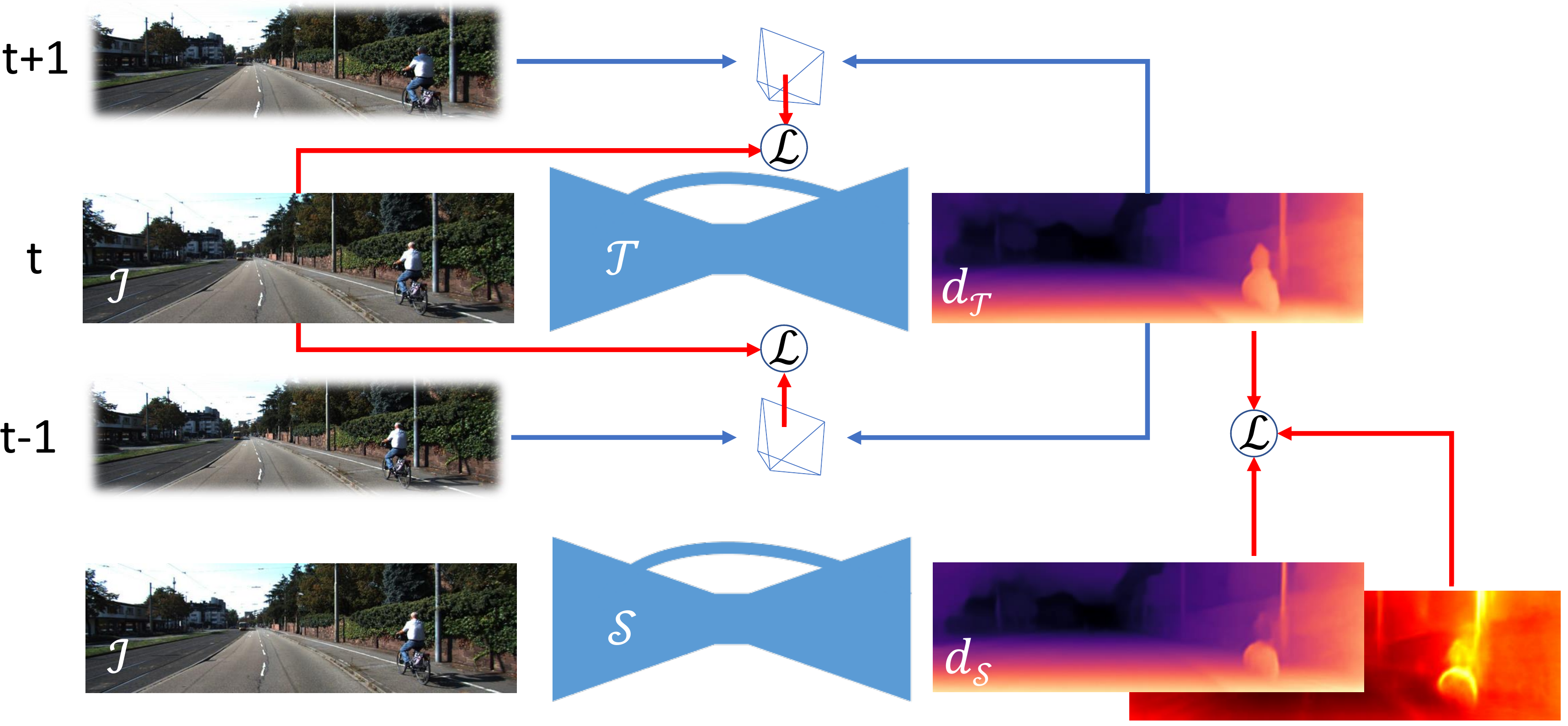}
    \caption{\textbf{Self-Teaching scheme.} A network $\mathcal{T}$ is trained in self-supervised fashion, \eg on monocular sequences $[t-1,t,t+1]$. A new instance $\mathcal{S}$ of the same is trained on $d_\mathcal{T}$ output of $\mathcal{T}$.}
    \label{fig:self-teaching}
\end{figure}

\textbf{Self-Teaching (\textit{Self}).} 
In order to decouple depth and pose when modelling uncertainty, we propose to source a direct form of supervision from the learned model itself. 
By training a first network in a self-supervised manner, we obtain a network instance $\mathcal{T}$ producing a noisy distribution $d_\mathcal{T}$. Then, we train a second instance of the same model, namely $\mathcal{S}$, to mimic the distribution sourced from $\mathcal{T}$. 
Typically, teacher-student frameworks \cite{zheng2019smooth} applied to monocular depth estimation \cite{Pilzer_2019_CVPR} deploy a complex architecture to supervise a more compact one. In contrast, in our approach the teacher $\mathcal{T}$ and the student $\mathcal{S}$ share the same architecture and for this reason we refer to it as Self-Teaching (\textit{Self}).
By assuming an L1 loss, we can model for instance negative log-likelihood minimization as

\begin{equation}
    \mathcal{L}_\text{\textit{Self}} = \frac{|\mu(d_\mathcal{S}) - d_\mathcal{T}|}{\sigma(d_\mathcal{S})} + \log \sigma(d_\mathcal{S})
\end{equation}
We will show how with this strategy i) we obtain a network $\mathcal{S}$ more accurate than $\mathcal{T}$ and ii) in case of monocular supervision, we can decouple depth from pose and achieve a much more effective uncertainty estimation.
Figure \ref{fig:self-teaching} summarizes our proposal.

\subsection{Bayesian estimation}

Finally, in Bayesian deep learning \cite{kendall2017uncertainties}, the model uncertainty can be explained by marginalizing over all possible $w$ rather than choosing a point estimate. According to Neal \cite{neal2012bayesian}, an approximate solution can be obtained by sampling N models and by modelling mean and variance as 

\begin{equation}
    p(d^*|\mathcal{I}, \mathcal{D}) \approx \sum_{i=1}^N p(d^*|\Theta(\mathcal{I},w_i))
\end{equation}
If mean and variance are modelled for each $w_i$ sampling, we can obtain overall mean and variance as reported in \cite{kendall2017uncertainties,ilg2018uncertainty}

\begin{equation}
    \mu(d) = \frac{1}{N} \sum_{i=1}^N \mu_i(d_i)
\end{equation}

\begin{equation}
    \sigma^2(d) = \frac{1}{N} \sum_{i=1}^N (\mu_i(d_i) - \mu(d))^2 + \sigma_i^2(d_i)
\end{equation}
The implementation of this approximation is straightforward by combining empirical and predictive methods \cite{kendall2017uncertainties,ilg2018uncertainty}. Purposely, in our experiments we will pick the best empirical and predictive methods, \eg combining \textit{Boot} and \textit{Self} (\textit{Boot+Self}).

\section{Experimental results}

In this section, we exhaustively evaluate self-supervised strategies for joint depth and uncertainty estimation. 

\subsection{Evaluation protocol, dataset and metrics}

At first, we describe all details concerning training and evaluation to ensure full reproducibility. Source code will be available at \url{https://github.com/mattpoggi/mono-uncertainty}. 

\textbf{Architecture and training schedule.} We choose as baseline model Monodepth2 \cite{Godard_2019_ICCV}, thanks to the code made available and to its possibility to be trained seamlessly according to monocular, stereo, or both self-supervision paradigms. 
In our experiments, we train any variant of this method following the protocol defined in \cite{Godard_2019_ICCV}, on batches of 12 images resized to $192\times640$ for 20 epochs starting from pre-trained encoders on ImageNet \cite{deng2009imagenet}. Moreover, we always follow the augmentation and training practices described in \cite{Godard_2019_ICCV}. Finally, to evaluate {\textit{Post}} we use the same weights made publicly available by the authors.
Regarding empirical methods, we set $N$ to 8 and the number of cycles $C$ for \textit{Snap} to 20. We randomly extract 25\% of the training set for each independent network in \textit{Boot}. Dropout is applied after convolutions in the decoder only. About predictive models, a single output channel is added in parallel to depth prediction channel.

\textbf{Dataset.} We compare all the models on the KITTI dataset \cite{geiger2013vision}, made of 61 scenes (about 42K stereo frames) acquired in driving scenarios. The dataset contains images at an average resolution of $375\times1242$ and depth maps from a calibrated LiDAR sensor. Following standards in the field, we deploy the Eigen split \cite{eigen2014depth} and set 80 meters as the maximum depth. For this purpose, we use the improved ground truth introduced in \cite{uhrig2017sparsity}, much more accurate than the raw LiDAR data, since our aim is a strict evaluation rather than a comparison with existing monocular methods.
Nevertheless, we report results on the raw LiDAR data using Garg’s crop \cite{garg2016unsupervised} as well in the supplementary material.

\textbf{Depth metrics.} To assess depth accuracy, we report for the sake of page limit three out of seven standard metrics\footnote{Results for the seven metrics are available as supplementary material} defined in \cite{eigen2014depth}. 
Specifically, we report the absolute relative error (Abs Rel), root mean square error (RMSE), and the amount of inliers ($\delta<1.25$). We refer the reader to \cite{eigen2014depth} or supplementary material for a complete description of these metrics. They enable a compact evaluation concerning both relative (Abs Rel and $\delta <1.25$) and absolute (RMSE) errors. Moreover, we also report the number of training iterations (\#Trn), parameters (\#Par), and forwards (\#Fwd) required at testing time to estimate depth. In the case of monocular supervision, we scale depth as in \cite{zhou2017unsupervised}.

\textbf{Uncertainty metrics.} To evaluate how significant the modelled uncertainties are, we use sparsification plots as in \cite{ilg2018uncertainty}. Given an error metric $\epsilon$, we sort all pixels in each depth map in order of descending uncertainty. Then, we iteratively extract a subset of pixels (\ie, 2\% in our experiments) and compute $\epsilon$ on the remaining to plot a curve, that is supposed to shrink if the uncertainty properly encodes the errors in the depth map. An ideal sparsification (\textit{oracle}) is obtained by sorting pixels in descending order of the $\epsilon$ magnitude. In contrast, a random uncertainty can be modelled as a constant, giving no information about how to remove erroneous measurements and, thus, a flat curve. By plotting the difference between estimated and oracle sparsification, we can measure the Area Under the Sparsification Error (AUSE, the \textbf{\textcolor{col1}{lower}} the better). Subtracting estimated sparsification from random one enables computing the Area Under the Random Gain (AURG, the \textbf{\textcolor{col2}{higher}} the better). The former quantifies how close the estimate is to the oracle uncertainty, the latter how better (or worse, as we will see in some cases) it is compared to no modelling at all.
We assume Abs Rel, RMSE or $\delta \ge 1.25$ (since $\delta < 1.25$ defines an accuracy score) as $\epsilon$.

\begin{table}
\centering
\scalebox{0.63}{
\renewcommand{\tabcolsep}{5pt} 
\begin{tabular}{c}
\begin{tabular}{l|c|r|r|r|cc|c}
    Method & Sup & \cellcolor{col1} \#Trn & \cellcolor{col1} \#Par & \cellcolor{col1} \#Fwd & \cellcolor{col1} Abs Rel & \cellcolor{col1} RMSE & \cellcolor{col2}$\delta<$1.25 \\
     \hline \hline
    Monodepth2 \cite{Godard_2019_ICCV} & M & $1\times$ & $1\times$ & $1\times$ & 0.090 & 3.942 & 0.914 \\
    Monodepth2-\textit{Post} \cite{Godard_2019_ICCV} & M & $1\times$ & $1\times$ & $2\times$ & 0.088 & 3.841 & 0.917 \\ \hline
    \hline
    Monodepth2-\textit{Drop} & M & $1\times$ & $1\times$ & N$\times$ & 0.101 & 4.146 & 0.892 \\
    Monodepth2-\textit{Boot} & M & N$\times$ & N$\times$ & $1\times$ & 0.092 & 3.821 & 0.911 \\
    Monodepth2-\textit{Snap} & M & 1$\times$ & N$\times$ & $1\times$ & 0.091 & 3.921 & 0.912 \\
    \hline
    Monodepth2-\textit{Repr} & M  & $1\times$ & $1\times$ & $1\times$ & 0.092 & 3.936 & 0.912 \\
    Monodepth2-\textit{Log} & M & $1\times$ & $1\times$ & $1\times$ & 0.091 & 4.052 & 0.910 \\
    Monodepth2-\textit{Self} & M & (1+1)$\times$ & $1\times$ & $1\times$ & \bfseries0.087 & 3.826 & \bfseries0.920 \\
    \hline
    Monodepth2-\textit{Boot+Log} & M & N$\times$ & N$\times$ & $1\times$ & 0.092 & 3.850 & 0.910 \\
    Monodepth2-\textit{Boot+Self} & M & (1+N)$\times$ & N$\times$ & $1\times$ & 0.088 & \bfseries 3.799 & 0.918 \\
    Monodepth2-\textit{Snap+Log} & M & $1\times$ & $1\times$ & $1\times$ & 0.092 & 3.961  & 0.911 \\    
    Monodepth2-\textit{Snap+Self} & M & (1+1)$\times$ & 1$\times$ & $1\times$ & 0.088 & 3.832 & 0.919 \\
    \hline \hline
\end{tabular}
\\
a) Depth evaluation
\\
\begin{tabular}{l| rr | rr | rr}
    & \multicolumn{2}{c|}{Abs Rel} & \multicolumn{2}{c|}{RMSE} & \multicolumn{2}{c}{$\delta \ge 1.25$} \\
    \hline
    Method & \cellcolor{col1} AUSE & \cellcolor{col2} AURG & \cellcolor{col1} AUSE & \cellcolor{col2} AURG & \cellcolor{col1} AUSE & \cellcolor{col2} AURG \\
    \hline \hline
    Monodepth2-\textit{Post} &   0.044 &   0.012 &   2.864 &   0.412 &   0.056 &   0.022 \\
    \hline
    Monodepth2-\textit{Drop} &   0.065 &   0.000 &   2.568 &   0.944 &   0.097 &   0.002 \\
    Monodepth2-\textit{Boot} &   0.058 &   0.001 &   3.982 &  -0.743 &   0.084 &  -0.001 \\
    Monodepth2-\textit{Snap} &   0.059 &  -0.001 &   3.979 &  -0.639 &   0.083 &  -0.002 \\
    \hline
    Monodepth2-\textit{Repr} &   0.051 &   0.008 &   2.972 &   0.381 &   0.069 &   0.013 \\
    Monodepth2-\textit{Log} &   0.039 &   0.020 &   2.562 &   0.916 &   0.044 &   0.038 \\
    Monodepth2-\textit{Self} & 0.030 &   0.026 &   2.009 &   1.266 &   0.030 &   0.045 \\
    \hline
    Monodepth2-\textit{Boot+Log} & 0.038 &   0.021 &   2.449 &   0.820 &   0.046 &   0.037  \\
    Monodepth2-\textit{Boot+Self} & \bfseries 0.029 &   \bfseries 0.028 &   \bfseries 1.924 &   \bfseries 1.316 &   \bfseries 0.028 &   \bfseries 0.049 \\
    Monodepth2-\textit{Snap+Log} & 0.038 &   0.022 &   2.385 &   1.001 &   0.043 &   0.039  \\
    Monodepth2-\textit{Snap+Self} & 0.031 &   0.026 &   2.043 &   1.230 &   0.030 &   0.045  \\
    \hline \hline
\end{tabular}
\\
b) Uncertainty evaluation 
\\
\end{tabular}
}
\caption{\textbf{Quantitative results for monocular (M) supervision.} Evaluation on Eigen split \cite{eigen2014depth} with improved ground truth \cite{uhrig2017sparsity}.}
\label{tab:mono}
\end{table}

\subsection{Monocular (M) supervision}

\textbf{Depth.} Table \ref{tab:mono}\textcolor{red}{a} reports depth accuracy for Monodepth2 variants implementing the different uncertainty estimation strategies when trained with monocular supervision. 
We can notice how, in general, empirical methods fail at improving depth prediction on most metrics, with \textit{Drop} having a large gap from the baseline. On the other hand, \textit{Boot} and \textit{Snap} slightly reduce RMSE.
Predictive methods as well produce worse depth estimates, except the proposed \textit{Self} method, which improves all the metrics compared to the baseline, even when post-processed.
Regarding the Bayesian solutions, both \textit{Boot} and \textit{Snap} performs worse when combined with \textit{Log}, while they are always improved by the proposed \textit{Self} method.

\textbf{Uncertainty.} Table \ref{tab:mono}\textcolor{red}{b} resumes performance of modelled uncertainties at reducing errors on the estimated depth maps. Surprisingly, empirical methods rarely perform better than the \textit{Post} solution. 
In particular, empirical methods alone fail at performing better than a random chance, except for \textit{Drop} that, on the other hand, produces much worse depth maps.
Predictive methods perform better, with \textit{Log} and \textit{Self} yielding the best results. Among them, our method outperforms \textit{Log} by a notable margin.
Combining empirical and predictive methods is beneficial, often improving over single choices. In particular, \textit{Boot+Self} achieves the best overall results.

\textbf{Summary.} In general \textit{Self}, combined with empirical methods, performs better for both depth accuracy and uncertainty modelling when dealing with M supervision, thanks to disentanglement between depth and pose. We believe that empirical methods performance can be ascribed to depth scale, being unknown during training.

\begin{table}
\centering
\scalebox{0.63}{
\begin{tabular}{c}
\begin{tabular}{l|c|r|r|r|cc|c}
    Method & Sup &  \cellcolor{col1} \#Trn & \cellcolor{col1} \#Par & \cellcolor{col1} \#Fwd & \cellcolor{col1} Abs Rel & \cellcolor{col1} RMSE & \cellcolor{col2}$\delta<$1.25 \\
     \hline \hline
    Monodepth2 \cite{Godard_2019_ICCV} & S & $1\times$ & $1\times$ & $1\times$ & 0.085  & 3.942  & 0.912  \\
    Monodepth2-\textit{Post} \cite{Godard_2019_ICCV} & S & $1\times$ & $1\times$ & $2\times$ & 0.084 & 3.777 & \bfseries0.915 \\ 
    \hline
    Monodepth2-\textit{Drop} & S & $1\times$ & $1\times$ & N$\times$ & 0.129 & 4.908 & 0.819 \\
    Monodepth2-\textit{Boot} & S & N$\times$ & N$\times$ & $1\times$ & 0.085 & \bfseries3.772 & 0.914  \\
    Monodepth2-\textit{Snap} & S & $1\times$ & N$\times$ & $1\times$ & 0.085 & 3.849 & 0.912  \\
    \hline
    Monodepth2-\textit{Repr} & S & $1\times$ & $1\times$ & $1\times$ & 0.085 & 3.873 & 0.913  \\
    Monodepth2-\textit{Log} & S & $1\times$ & $1\times$ & $1\times$ & 0.085 & 3.860 & \bfseries0.915  \\
    Monodepth2-\textit{Self} & S & (1+1)$\times$ & $1\times$ & $1\times$ & 0.084 & 3.835 & \bfseries0.915  \\
    \hline
    Monodepth2-\textit{Boot+Log} & S & N$\times$ & N$\times$ & $1\times$ & 0.085 & 3.777 & 0.913  \\    
    Monodepth2-\textit{Boot+Self} & S & (1+N)$\times$ & N$\times$ & $1\times$ & 0.085 & 3.793 & 0.914 \\
    Monodepth2-\textit{Snap+Log} & S & 1$\times$ & 1$\times$ & $1\times$ & \bfseries0.083 & 3.833 & 0.914  \\    
    Monodepth2-\textit{Snap+Self} & S & (1+1)$\times$ & 1$\times$ & $1\times$ & 0.086 & 3.859 & 0.912 \\    
    \hline \hline
\end{tabular}
\\
a) Depth evaluation
\\
\begin{tabular}{l| rr | rr | rr}
    & \multicolumn{2}{c|}{Abs Rel} & \multicolumn{2}{c|}{RMSE} & \multicolumn{2}{c}{$\delta \ge 1.25$} \\
    \hline
    Method & \cellcolor{col1} AUSE & \cellcolor{col2} AURG & \cellcolor{col1} AUSE & \cellcolor{col2} AURG & \cellcolor{col1} AUSE & \cellcolor{col2} AURG \\
    \hline \hline
    Monodepth2-\textit{Post} &   0.036 &   0.020 &   2.523 &   0.736 &   0.044 &   0.034  \\
    \hline
    Monodepth2-\textit{Drop} &   0.103 &  -0.029 &   6.163 &  -2.169 &   0.231 &  -0.080 \\
    Monodepth2-\textit{Boot} &   0.028 &   0.029 &   2.291 &   0.964 &   0.031 &   0.048 \\
    Monodepth2-\textit{Snap} &   0.028 &   0.029 &   2.252 &   1.077 &   0.030 &   0.051 \\
    \hline
    Monodepth2-\textit{Repr} &   0.040 &   0.017 &   2.275 &   1.074 &   0.050 &   0.030 \\
    Monodepth2-\textit{Log} &   0.022 &   0.036 &   0.938 &   2.402 &   \bfseries 0.018 &   0.061 \\
    Monodepth2-\textit{Self} &   0.022 &   0.035 &   1.679 &   1.642 &   0.022 &   0.056 \\
    \hline
    Monodepth2-\textit{Boot+Log} &   \bfseries 0.020 &   \bfseries 0.038 &   \bfseries 0.807 &   \bfseries 2.455 &   \bfseries 0.018 &   \bfseries 0.063 \\
    Monodepth2-\textit{Boot+Self} &   0.023 &   0.035 &   1.646 &   1.628 &   0.021 &   0.058 \\
    Monodepth2-\textit{Snap+Log} &   0.021 &   0.037 &   0.891 &   2.426 &   \bfseries 0.018 &   0.061 \\
    Monodepth2-\textit{Snap+Self} &   0.023 &   0.035 &   1.710 &   1.623 &   0.023 &   0.058 \\
    \hline \hline
\end{tabular}
\\
b) Uncertainty evaluation 
\\
\end{tabular}
}
\caption{\textbf{Quantitative results for stereo (S) supervision.} Evaluation on Eigen split \cite{eigen2014depth} with improved ground truth \cite{uhrig2017sparsity}.}
\label{tab:stereo}
\end{table}

\subsection{Stereo (S) supervision}

\textbf{Depth.} On Table \ref{tab:stereo}\textcolor{red}{a} we show the results of the same approaches when trained with stereo supervision. Again, \textit{Drop} fails to improve depth accuracy, together with \textit{Repr} among predictive methods. \textit{Boot} produces the best improvement, in particular in terms of RMSE. Traditional \textit{Log} improves this time over the baseline, according to RMSE and $\delta <1.25$ metrics while, \textit{Self} consistently improves the baseline on all metrics, although it does not outperform \textit{Post}, which requires two forward passes. 

\textbf{Uncertainty.} Table \ref{tab:stereo}\textcolor{red}{b} summarizes the effectiveness of modelled uncertainties. This time, only \textit{Drop} performs worse than \textit{Post} achieving negative AURG, thus being detrimental at sparsification, while other empirical methods achieve much better results.
In these experiments, thanks to the known pose of the stereo setup, \textit{Log} deals only with depth uncertainty and thus performs extremely well. \textit{Self}, although allowing for more accurate depth as reported in Table \ref{tab:stereo}\textcolor{red}{a}, ranks second this time. Considering Bayesian implementations, again, both \textit{Boot} and \textit{Snap} are always improved. Conversely, compared to the M case, \textit{Log} this time consistently outperforms \textit{Self} in any Bayesian formulation.

\textbf{Summary.} When the pose is known, the gap between \textit{Log} and \textit{Self} concerning depth accuracy is minor, with \textit{Self} performing better when modelling only predictive uncertainty and \textit{Log} slightly better with Bayesian formulations. For uncertainty estimation, \textit{Log} consistently performs better.
The behavior of empirical methods alone confirms our findings from the previous experiments: by knowing the scale, \textit{Boot} and \textit{Snap} model uncertainty much better. In contrast, \textit{Drop} fails for this purpose.

\begin{table}
\centering
\scalebox{0.63}{
\begin{tabular}{c}

\begin{tabular}{l|c|r|r|r|cc|c}
    Method & Sup & \cellcolor{col1} \#Trn & \cellcolor{col1} \#Par & \cellcolor{col1} \#Fwd & \cellcolor{col1} Abs Rel & \cellcolor{col1} RMSE & \cellcolor{col2}$\delta<$1.25 \\
     \hline \hline
    Monodepth2 \cite{Godard_2019_ICCV} & MS & $1\times$ & $1\times$ & $1\times$ & 0.084 & 3.739 & 0.918 \\
    Monodepth2-\textit{Post} \cite{Godard_2019_ICCV} & MS & $1\times$ & $1\times$ & $2\times$ & \bfseries0.082 & \bfseries3.666 & \bfseries0.919 \\ 
    \hline
    Monodepth2-\textit{Drop} & MS & $1\times$ & 1$\times$ & N$\times$ & 0.172 & 5.885 & 0.679 \\
    Monodepth2-\textit{Boot} & MS & N$\times$ & N$\times$ & 1$\times$ & 0.086 & 3.787 & 0.910 \\
    Monodepth2-\textit{Snap} & MS & $1\times$ & N$\times$ & $1\times$ & 0.085 & 3.806 & 0.914 \\
    \hline
    Monodepth2-\textit{Repr} & MS & $1\times$ & $1\times$ & $1\times$ & 0.084 & 3.828 & 0.913 \\
    Monodepth2-\textit{Log} & MS & $1\times$ & $1\times$ & $1\times$ & 0.083 & 3.790 & 0.916 \\
    Monodepth2-\textit{Self} & MS & (1+1)$\times$ & $1\times$ & $1\times$ & 0.083 & 3.682 & \bfseries0.919 \\
    \hline
    Monodepth2-\textit{Boot+Log}& MS & N$\times$ & N$\times$ & 1$\times$ & 0.086 & 3.771 & 0.911\\
    Monodepth2-\textit{Boot+Self} & MS & (1+N)$\times$ & N$\times$ & 1$\times$ & 0.085 & 3.704 & 0.915 \\
    Monodepth2-\textit{Snap+Log} & MS & 1$\times$ & 1$\times$ & 1$\times$ & 0.084 & 3.828 & 0.914 \\  
    Monodepth2-\textit{Snap+Self} & MS & (1+1)$\times$ & 1$\times$ & 1$\times$ & 0.085 & 3.715 & 0.916 \\    
    \hline \hline
\end{tabular}
\\
a) Depth evaluation
\\
\begin{tabular}{l| rr | rr | rr}
    & \multicolumn{2}{c|}{Abs Rel} & \multicolumn{2}{c|}{RMSE} & \multicolumn{2}{c}{$\delta \ge 1.25$} \\
    \hline
    Method & \cellcolor{col1} AUSE & \cellcolor{col2} AURG & \cellcolor{col1} AUSE & \cellcolor{col2} AURG & \cellcolor{col1} AUSE & \cellcolor{col2} AURG \\
    \hline \hline
    Monodepth2-\textit{Post} &   0.036 &   0.018 &   2.498 &   0.655 &   0.044 &   0.031 \\
    \hline
    Monodepth2-\textit{Drop} &   0.103 &  -0.027 &   7.114 &  -2.580 &   0.303 &  -0.081 \\
    Monodepth2-\textit{Boot} &   0.028 &   0.030 &   2.269 &   0.985 &   0.034 &   0.049 \\
    Monodepth2-\textit{Snap} &   0.029 &   0.028 &   2.245 &   1.029 &   0.033 &   0.047 \\
    \hline
    Monodepth2-\textit{Repr} &   0.046 &   0.010 &   2.662 &   0.635 &   0.062 &   0.018 \\
    Monodepth2-\textit{Log} &   0.028 &   0.029 &   1.714 &   \bfseries1.562 &   0.028 &   0.050 \\
    Monodepth2-\textit{Self} &   \bfseries0.022 &   0.033 &   \bfseries1.654 &  1.515 &   \bfseries0.023 &   0.052 \\
    \hline
    Monodepth2-\textit{Boot+Log} &   0.030 &   0.028 &   1.962 &   1.282 &   0.032 &   0.051 \\
    Monodepth2-\textit{Boot+Self} &   0.023 &   0.033 &   1.688 &   1.494 &   \bfseries0.023 &   \bfseries0.056 \\
    Monodepth2-\textit{Snap+Log} &   0.030 &   0.027 &   2.032 &   1.272 &   0.032 &   0.048 \\
    Monodepth2-\textit{Snap+Self} &   0.023 &   \bfseries0.034 &   1.684 &   1.510 &   \bfseries0.023 &   0.055 \\    
    \hline \hline
\end{tabular}
\\
b) Uncertainty evaluation 
\\
\end{tabular}
}
\caption{\textbf{Quantitative results for monocular+stereo (MS) supervision.} Evaluation on Eigen split \cite{eigen2014depth} with improved ground truth \cite{uhrig2017sparsity}.}
\label{tab:mono+stereo}
\end{table}

\subsection{Monocular+Stereo (MS) supervision}

\textbf{Depth.} Table \ref{tab:mono+stereo}\textcolor{red}{a} reports the behavior of depth accuracy when monocular and stereo supervisions are combined. 
In this case, only \textit{Self} consistently outperforms the baseline and is competitive with \textit{Post}, which still requires two forward passes. Among empirical methods, \textit{Boot} is the most effective. Regarding Bayesian solutions, those using \textit{Self} are, in general, more accurate on most metrics, yet surprisingly worse than \textit{Self} alone.

\textbf{Uncertainty.} Table \ref{tab:mono+stereo}\textcolor{red}{b} shows the performance of the considered uncertainties. The behavior of all variants is similar to the one observed with stereo supervision, except for \textit{Log} and \textit{Self}. We can notice that \textit{Self} outperforms \textit{Log}, similarly to what observed with M supervision. It confirms that pose estimation drives \textit{Log} to worse uncertainty estimation, while \textit{Self} models are much better thanks to the training on proxy labels produced by the Teacher network.
Concerning Bayesian solutions, in general, \textit{Boot} and \textit{Snap} are improved when combined with both \textit{Log} and \textit{Self}, with \textit{Self} combinations typically better than their \textit{Log} counterparts and equivalent to standalone \textit{Self}. 

\textbf{Summary.} The evaluation with monocular and stereo supervision confirms that when the pose is estimated alongside with depth, \textit{Self} proves to be a better solution compared to \textit{Log} and, in general, other approaches to model uncertainty.
Finally, empirical methods alone behave as for experiments with stereo supervision, confirming that the knowledge of the scale during training is crucial to the proper behavior of \textit{Drop}, \textit{Boot} and \textit{Snap}.

\subsection{Sparsification curves}

\begin{figure}
    \centering
    \renewcommand{\tabcolsep}{0.1pt} 
    \begin{tabular}{cccccc}
        \includegraphics[height=0.18\textwidth]{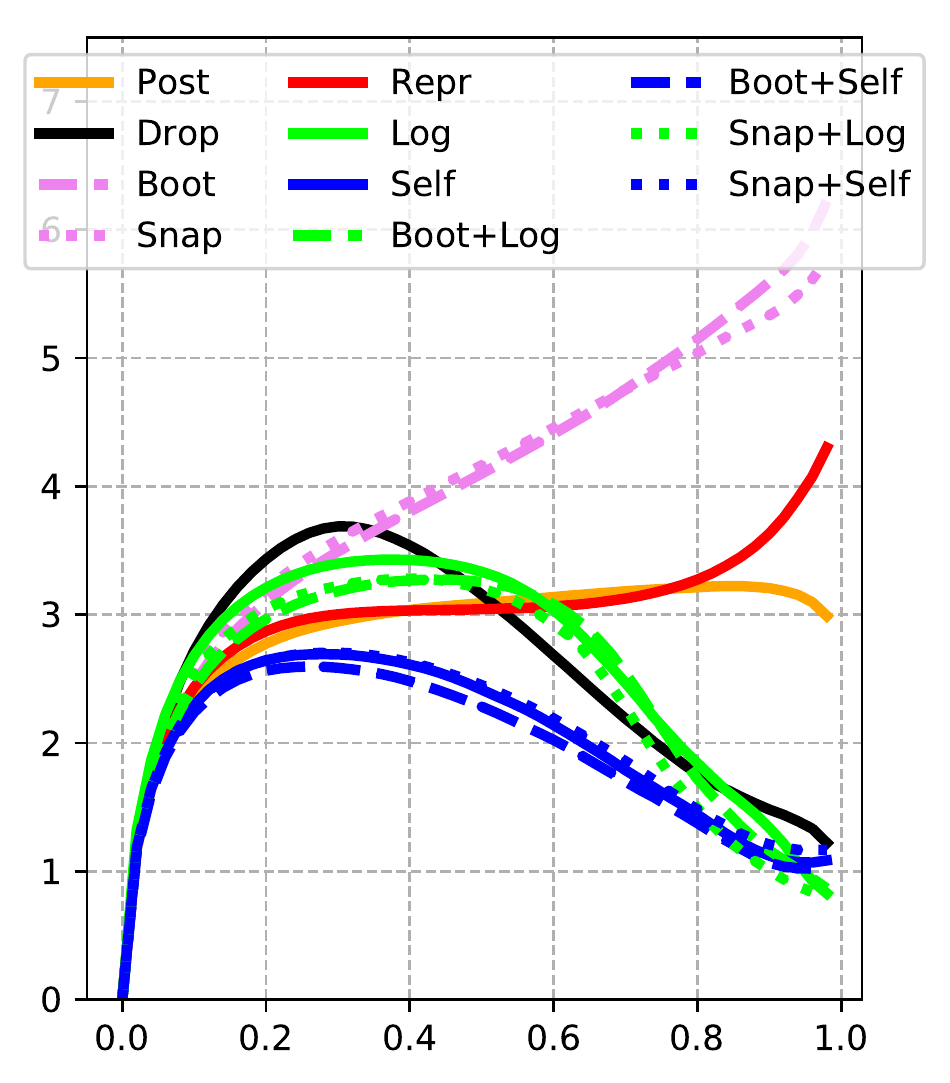}  &
        \includegraphics[height=0.18\textwidth]{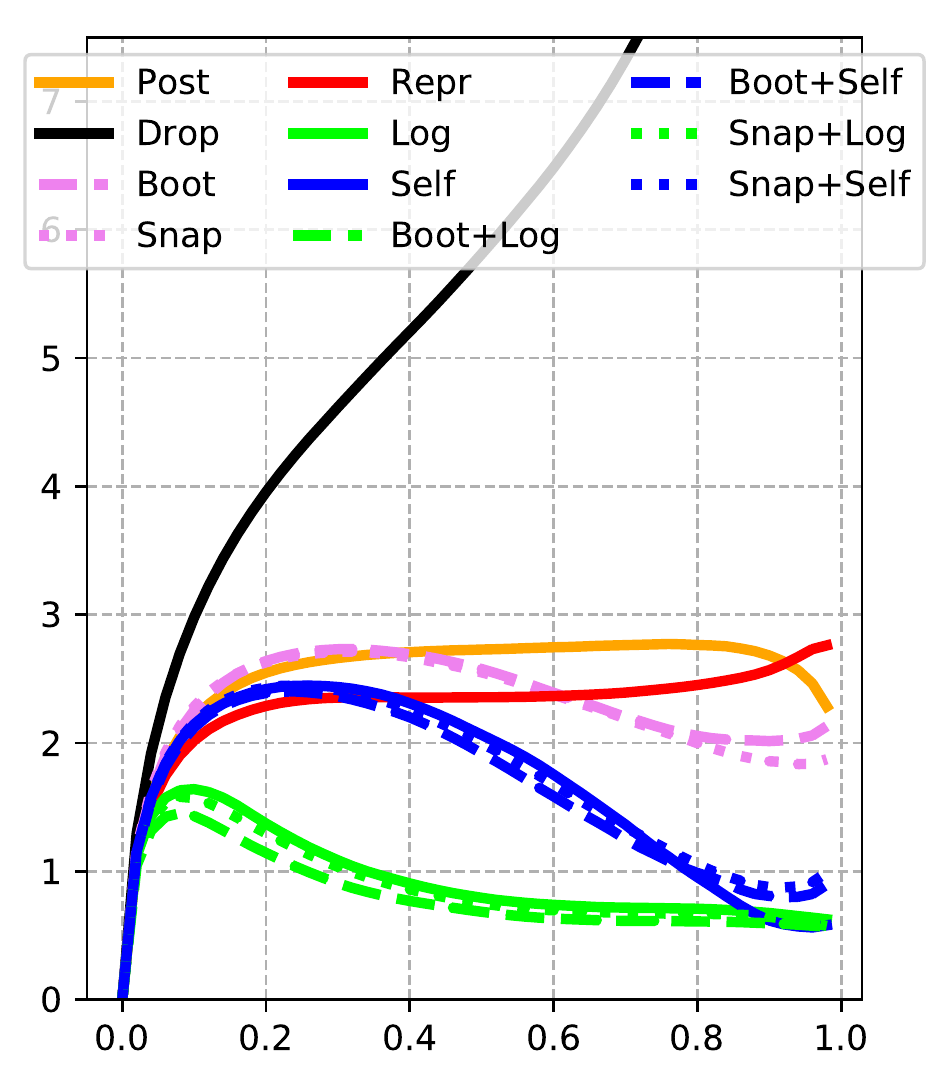} &
        \includegraphics[height=0.18\textwidth]{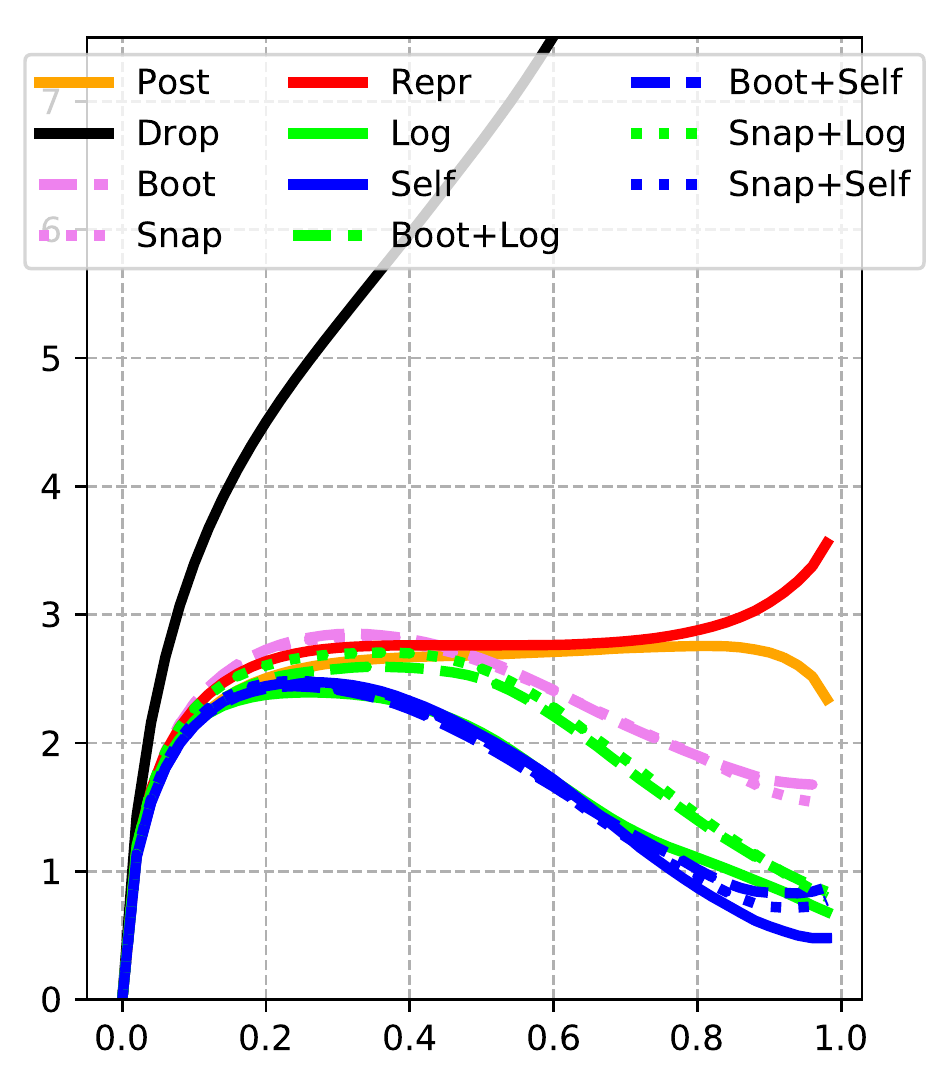} \\
    \end{tabular}
    \caption{\textbf{Sparsification Error curves.} From left to right, average RMSE with M, S and MS supervisions. Best viewed with colors.}
    \label{fig:visual_quantitatives}
\end{figure}

In order to further outline our findings, we report in Figure \ref{fig:visual_quantitatives} the RMSE sparsification error curves, averaged over the test set, when training with M, S or MS supervision. The plots show that methods leveraging on \textit{Self} (blue) are the best to model uncertainty when dealing with pose estimation, \ie M and MS, while those using \textit{Log} (green) are better when training on S. We report curves for Abs Rel and $\delta \ge 1.25$ in the supplementary material. 

\subsection{Supplementary material}

For the sake of the pages limit, we report more details about the experiments shown so far in the supplementary material. Specifically, i) complete depth evaluation with all seven metrics defined in \cite{eigen2014depth}, ii) depth and uncertainty evaluation with reduced depth range to 50 meters, iii) evaluation assuming the raw LiDAR data as ground truth, for compliancy with previous works \cite{Godard_2019_ICCV} and iv) sparsification curves for all metrics. We also provide additional qualitative results in the form of images and a video sequence, available at \url{www.youtube.com/watch?v=bxVPXqf4zt4}.

\section{Conclusion}

In this paper, we have thoroughly investigated for the first time in literature uncertainty modelling in self-supervised monocular depth estimation. We have reviewed and evaluated existing techniques, as well as introduced a novel Self-Teaching (\textit{Self}) paradigm. We have considered up to 11 strategies to estimate the uncertainty on predictions of a depth-from-mono network trained in a self-supervised manner.
Our experiments highlight how different supervision strategies lead to different winners among the considered methods. In particular, among empirical methods, only Dropout sampling performs well when the scale is unknown during training (M), while it is the only one failing when scale is known (S, MS).
Empirical methods are affected by pose estimation, for which log-likelihood maximization gives sub-optimal results when the pose is unknown (M, MS). In these latter cases, potentially the most appealing for practical applications, the proposed \textit{Self} technique results in the best strategy to model uncertainty. Moreover, uncertainty estimation also improves depth accuracy consistently, with any training paradigm.

\textbf{Acknowledgement.} We gratefully acknowledge the support of NVIDIA Corporation with the donation of the Titan Xp GPU used for this research. 

{\small
\bibliographystyle{ieee_fullname}
\bibliography{ref}
}

\newpage\phantom{Supplementary}
\multido{\i=1+1}{15}{


\section*{Supplementary material}

# On the uncertainty of self-supervised monocular depth estimation – Supplementary material

Matteo Poggi   Filippo Aleotti   Fabio Tosi   Stefano Mattoccia

Department of Computer Science and Engineering (DISI)
University of Bologna, Italy

{m.poggi, filippo.aleotti2, fabio.tosi5, stefano.mattoccia }@unibo.it

In this document, we provide more detailed results concerning the experiments reported in the paper "On the uncertainty of self-supervised monocular depth estimation". As in the submitted paper, we often simplify the notation by referring to *self-supervision* as *supervision*.

## 1. Insights about sparsification over different metrics

In the paper, due to the lack of space, we choose to show sparsification performance over three metrics, respectively Abs Rel, RMSE and $\delta \geq 1.25$. The first two metrics concern with the sparsification of an average error over a single depth map. In other words, this explains how good is our uncertainty modelling at finding pixels with the highest errors in magnitude and thus how much we can reduce the overall average error by removing them accordingly. The difference between the two consists in the fact that Abs Rel is normalized over the ground truth depth value, *i.e.* the magnitude of the error decreases for points farther from the camera, while the RMSE is independent of the depth in the scene.

Differently, $\delta \geq 1.25$ metric selects a set of pixels (*i.e.* for which estimated depth is greater/smaller than $1.25\times$ the ground-truth value) as outliers. By sparsification, according to this metric, we aim at reducing the percentage of outliers in the depth map.

## 2. Detailed depth evaluation

In this document, we report the complete evaluation of each Monodepth2 variant on the seven metrics traditionally adopted in this field [2], obtained as follows:

$$\text{Abs Rel} = \frac{1}{||\mathcal{I}||} \sum_{p \in \mathcal{I}} \frac{|d(p) - d^*(p)|}{d^*(p)} \tag{1}$$

$$\text{Sq Rel} = \frac{1}{||\mathcal{I}||} \sum_{p \in \mathcal{I}} \frac{(d(p) - d^*(p))^2}{d^*} \tag{2}$$

$$\text{RMSE} = \sqrt{\frac{1}{||\mathcal{I}||} \sum_{p \in \mathcal{I}} (d(p) - d^*(p))^2} \tag{3}$$

$$\text{RMSE log} = \sqrt{\frac{1}{||\mathcal{I}||} \sum_{p \in \mathcal{I}} (\log d(p) - \log d^*(p))^2} \tag{4}$$

$$\delta < 1.25^k = \frac{1}{||\mathcal{I}||} \sum_{p \in \mathcal{I}} \max\left(\frac{d}{d^*}, \frac{d^*}{d}\right) < 1.25^k \tag{5}$$

with $d$, $d^*$ respectively estimated and ground truth depth maps, $p$ a single pixel from input image $\mathcal{I}$ and $||\mathcal{I}||$ the total amount of pixels in $\mathcal{I}$. Tables 1, 2 and 3 exhaustively collect results on the Eigen test split [2] using the improved ground truth made available in [6], respectively when using monocular (M), stereo (S) or both (MS) (self-)supervisions. Since the ground truth is not provided for all 697 images, we reduce this split to 652 according to previous works [1, 5, 7].

1

| Method | Sup | #Trn | #Par | #Fwd | Abs Rel | Sq Rel | RMSE | RMSE log | $\delta < 1.25$ | $\delta < 1.25^2$ | $\delta < 1.25^3$ |
|---|---|---|---|---|---|---|---|---|---|---|---|
| Monodepth2 [5] | M | 1× | 1× | 1× | 0.090 | 0.545 | 3.942 | 0.137 | 0.914 | **0.983** | **0.995** |
| Monodepth2-*Post* [5] | M | 1× | 1× | 2× | 0.088 | 0.508 | 3.843 | 0.134 | 0.917 | **0.983** | **0.995** |
| Monodepth2-*Drop* | M | 1× | 1× | N× | 0.101 | 0.596 | 4.148 | 0.150 | 0.892 | 0.976 | 0.994 |
| Monodepth2-*Boot* | M | N× | N× | 1× | 0.092 | **0.505** | 3.823 | 0.136 | 0.911 | 0.982 | **0.995** |
| Monodepth2-*Snap* | M | 1× | N× | 1× | 0.091 | 0.532 | 3.923 | 0.137 | 0.912 | **0.983** | **0.995** |
| Monodepth2-*Repr* | M | 1× | 1× | 1× | 0.092 | 0.543 | 3.936 | 0.138 | 0.912 | 0.981 | **0.995** |
| Monodepth2-*Log* | M | 1× | 1× | 1× | 0.091 | 0.588 | 4.053 | 0.139 | 0.911 | 0.980 | **0.995** |
| Monodepth2-*Self* | M | (1+1)× | 1× | 1× | **0.087** | 0.514 | 3.827 | **0.133** | **0.920** | **0.983** | **0.995** |
| Monodepth2-*Boot+Log* | M | N× | N× | 1× | 0.092 | 0.509 | 3.852 | 0.137 | 0.910 | 0.982 | **0.995** |
| Monodepth2-*Boot+Self* | M | (1+N)× | N× | 1× | 0.088 | 0.507 | **3.800** | **0.133** | 0.918 | **0.983** | **0.995** |
| Monodepth2-*Snap+Log* | M | 1× | 1× | 1× | 0.092 | 0.564 | 3.961 | 0.139 | 0.911 | 0.981 | 0.994 |
| Monodepth2-*Snap+Self* | M | (1+1)× | 1× | 1× | 0.088 | 0.518 | 3.833 | **0.133** | 0.919 | **0.983** | **0.995** |

Table 1. **Depth evaluation for monocular (M) supervision.** Evaluation on the Eigen test split [2] with improved ground truth [6].

| Method | Sup | #Trn | #Par | #Fwd | Abs Rel | Sq Rel | RMSE | RMSE log | $\delta < 1.25$ | $\delta < 1.25^2$ | $\delta < 1.25^3$ |
|---|---|---|---|---|---|---|---|---|---|---|---|
| Monodepth2 [5] | S | 1× | 1× | 1× | 0.085 | 0.537 | 3.942 | 0.139 | 0.912 | 0.979 | 0.993 |
| Monodepth2-*Post* [5] | S | 1× | 1× | 2× | **0.084** | 0.504 | 3.777 | 0.137 | **0.915** | 0.980 | **0.994** |
| Monodepth2-*Drop* | S | 1× | 1× | N× | 0.129 | 0.791 | 4.908 | 0.187 | 0.819 | 0.959 | 0.990 |
| Monodepth2-*Boot* | S | N× | N× | 1× | 0.085 | 0.511 | **3.772** | 0.137 | 0.914 | 0.980 | **0.994** |
| Monodepth2-*Snap* | S | 1× | N× | 1× | 0.085 | 0.535 | 3.849 | 0.139 | 0.912 | 0.980 | 0.993 |
| Monodepth2-*Repr* | S | 1× | 1× | 1× | 0.085 | 0.532 | 3.873 | 0.140 | 0.913 | 0.979 | 0.993 |
| Monodepth2-*Log* | S | 1× | 1× | 1× | 0.085 | 0.535 | 3.860 | 0.140 | **0.915** | 0.979 | 0.993 |
| Monodepth2-*Self* | S | (1+1)× | 1× | 1× | **0.084** | 0.524 | 3.835 | 0.137 | **0.915** | 0.980 | 0.993 |
| Monodepth2-*Boot+Log* | S | N× | N× | 1× | 0.085 | 0.511 | 3.777 | 0.137 | 0.913 | 0.980 | **0.994** |
| Monodepth2-*Boot+Self* | S | (1+N)× | N× | 1× | 0.085 | 0.510 | 3.792 | **0.135** | 0.914 | **0.981** | **0.994** |
| Monodepth2-*Snap+Log* | S | 1× | 1× | 1× | **0.084** | 0.529 | 3.833 | 0.138 | 0.914 | 0.980 | **0.994** |
| Monodepth2-*Snap+Self* | S | (1+1)× | 1× | 1× | 0.086 | 0.532 | 3.858 | 0.138 | 0.912 | 0.980 | **0.994** |

Table 2. **Depth evaluation for stereo (S) supervision.** Evaluation on the Eigen test split [2] with improved ground truth [6].

| Method | Sup | #Trn | #Par | #Fwd | Abs Rel | Sq Rel | RMSE | RMSE log | $\delta < 1.25$ | $\delta < 1.25^2$ | $\delta < 1.25^3$ |
|---|---|---|---|---|---|---|---|---|---|---|---|
| Monodepth2 [5] | MS | 1× | 1× | 1× | 0.084 | 0.494 | 3.739 | 0.132 | 0.918 | 0.983 | **0.995** |
| Monodepth2-*Post* [5] | MS | 1× | 1× | 2× | **0.082** | **0.470** | **3.666** | **0.129** | **0.919** | **0.984** | **0.995** |
| Monodepth2-*Drop* | MS | 1× | 1× | N× | 0.172 | 1.074 | 5.886 | 0.237 | 0.679 | 0.933 | 0.982 |
| Monodepth2-*Boot* | MS | N× | N× | 1× | 0.086 | 0.497 | 3.787 | 0.136 | 0.910 | 0.981 | **0.995** |
| Monodepth2-*Snap* | MS | 1× | N× | 1× | 0.085 | 0.504 | 3.803 | 0.134 | 0.914 | 0.983 | **0.995** |
| Monodepth2-*Repr* | MS | 1× | 1× | 1× | 0.084 | 0.500 | 3.829 | 0.134 | 0.913 | 0.982 | **0.995** |
| Monodepth2-*Log* | MS | 1× | 1× | 1× | 0.083 | 0.518 | 3.789 | 0.132 | 0.916 | **0.984** | **0.995** |
| Monodepth2-*Self* | MS | (1+1)× | 1× | 1× | 0.083 | 0.485 | 3.682 | 0.130 | **0.919** | **0.984** | **0.995** |
| Monodepth2-*Boot+Log* | MS | N× | N× | 1× | 0.086 | 0.497 | 3.771 | 0.135 | 0.911 | 0.981 | **0.995** |
| Monodepth2-*Boot+Self* | MS | (1+N)× | N× | 1× | 0.085 | 0.486 | 3.704 | 0.131 | 0.915 | 0.983 | **0.995** |
| Monodepth2-*Snap+Log* | MS | 1× | 1× | 1× | 0.084 | 0.512 | 3.828 | 0.134 | 0.914 | 0.982 | **0.995** |
| Monodepth2-*Snap+Self* | MS | (1+1)× | 1× | 1× | 0.085 | 0.497 | 3.714 | 0.131 | 0.916 | 0.983 | **0.995** |

Table 3. **Depth evaluation for monocular+stereo (MS) supervision.** Evaluation on the Eigen test split [2] with improved ground truth [6].

## 3. Sparsification curves

Figures 1, 2 and 3 show Sparsification Error curves computed for all the three metrics evaluated in the submitted paper, *i.e.* Abs Rel, RMSE and $\delta \geq 1.25$, respectively for M, S and MS supervisions. The curves highlight consistent behaviour on each metric, confirming that Self-Teaching strategies (blue) outperform traditional log-likelihood maximization (green) on M and MS, with these latter ones yielding better results with S supervision.

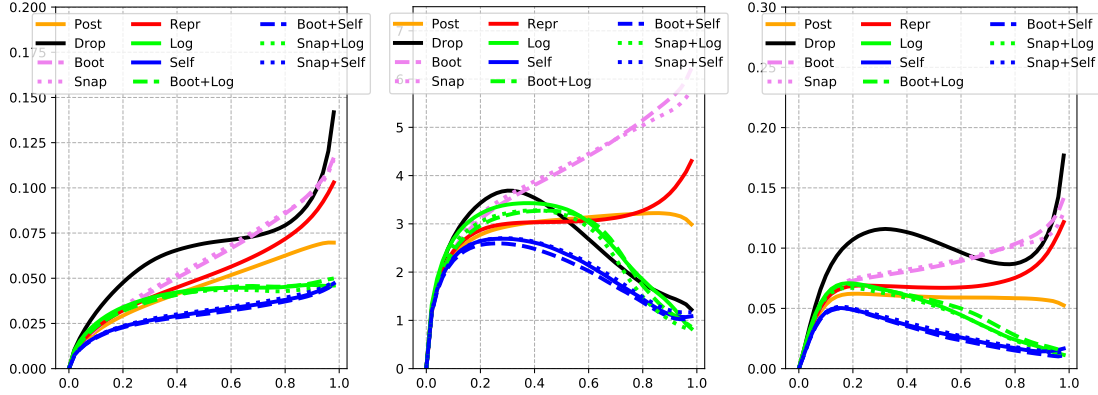

Figure 1. **Sparsification Error curves for monocular (M) supervision.** From left to right, Abs Rel, RMSE and $\delta \geq 1.25$.

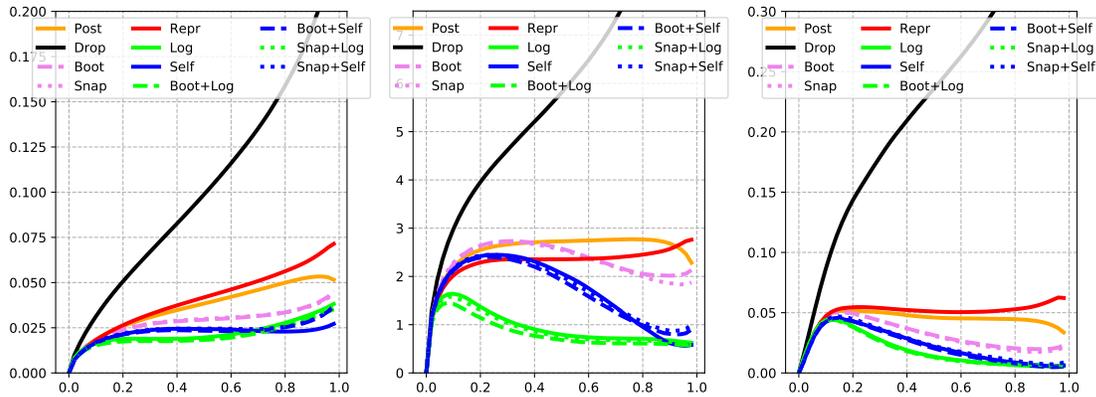

Figure 2. **Sparsification Error curves for stereo (S) supervision.** From left to right, Abs Rel, RMSE and $\delta \geq 1.25$.

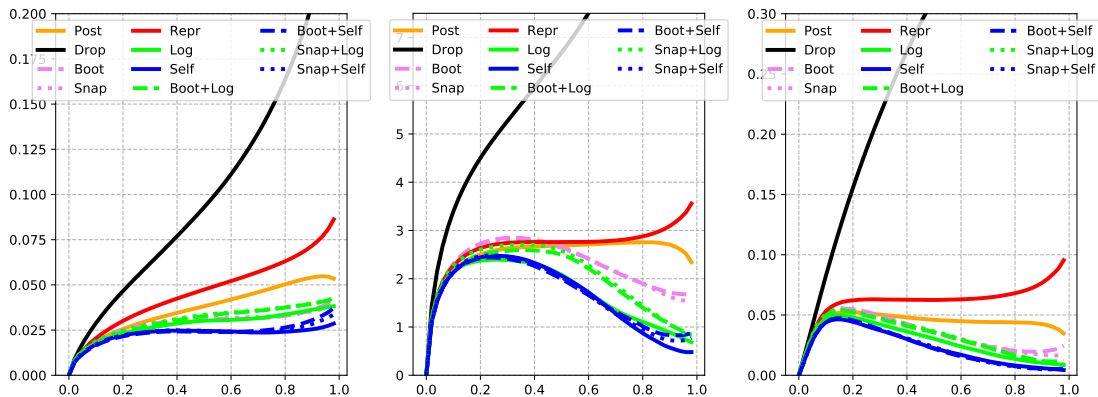

Figure 3. **Sparsification Error curves for monocular+stereo (MS) supervision.** From left to right, Abs Rel, RMSE and $\delta \geq 1.25$.

# 4. Depth evaluation – 50 meters cap

Consistently with previous works [3, 4], we also report results obtained by capping depth range to 50 meters. We can notice how, conversely to the evaluation carried out on raw LiDAR traditionally performed by existing works [3, 4], the margin between evaluating at 80 or 50 meters is much lower.

| Method | Sup | #Trn | #Par | #Fwd | Abs Rel | Sq Rel | RMSE | RMSE log | $\delta < 1.25$ | $\delta < 1.25^2$ | $\delta < 1.25^3$ |
|---|---|---|---|---|---|---|---|---|---|---|---|
| Monodepth2 [5] | M | 1× | 1× | 1× | 0.090 | 0.502 | 3.979 | 0.137 | 0.911 | 0.983 | 0.995 |
| Monodepth2-*Post* [5] | M | 1× | 1× | 2× | 0.088 | **0.476** | 3.914 | 0.134 | 0.914 | **0.984** | **0.996** |
| Monodepth2-*Drop* | M | 1× | 1× | N× | 0.101 | 0.580 | 4.239 | 0.151 | 0.889 | 0.976 | 0.994 |
| Monodepth2-*Boot* | M | N× | N× | 1× | 0.092 | 0.494 | 3.958 | 0.138 | 0.907 | 0.982 | **0.996** |
| Monodepth2-*Snap* | M | 1× | N× | 1× | 0.091 | 0.502 | 3.997 | 0.137 | 0.909 | 0.983 | **0.996** |
| Monodepth2-*Repr* | M | 1× | 1× | 1× | 0.091 | 0.504 | 3.991 | 0.138 | 0.909 | 0.982 | 0.995 |
| Monodepth2-*Log* | M | 1× | 1× | 1× | 0.091 | 0.588 | 4.053 | 0.139 | 0.911 | 0.980 | 0.995 |
| Monodepth2-*Self* | M | (1+1)× | 1× | 1× | **0.086** | 0.477 | 3.898 | **0.133** | **0.916** | 0.983 | 0.995 |
| Monodepth2-*Boot+Log* | M | N× | N× | 1× | 0.092 | 0.495 | 3.970 | 0.138 | 0.907 | 0.982 | 0.995 |
| Monodepth2-*Boot+Self* | M | (1+N)× | N× | 1× | 0.088 | 0.477 | **3.885** | **0.133** | 0.914 | 0.983 | **0.996** |
| Monodepth2-*Snap+Log* | M | 1× | 1× | 1× | 0.091 | 0.526 | 4.011 | 0.139 | 0.908 | 0.981 | 0.994 |
| Monodepth2-*Snap+Self* | M | (1+1)× | 1× | 1× | 0.088 | 0.478 | 3.890 | **0.133** | 0.915 | 0.983 | **0.996** |

Table 4. **Depth evaluation for monocular (M) supervision.** Evaluation on the Eigen test split [2] with improved ground truth [6]. Maximum depth reduced to 50 meters.

| Method | Sup | #Trn | #Par | #Fwd | Abs Rel | Sq Rel | RMSE | RMSE log | $\delta < 1.25$ | $\delta < 1.25^2$ | $\delta < 1.25^3$ |
|---|---|---|---|---|---|---|---|---|---|---|---|
| Monodepth2 [5] | S | 1× | 1× | 1× | 0.084 | 0.491 | 3.907 | 0.139 | 0.908 | 0.980 | 0.994 |
| Monodepth2-*Post* [5] | S | 1× | 1× | 2× | **0.083** | 0.470 | 3.848 | 0.137 | **0.911** | 0.981 | 0.994 |
| Monodepth2-*Drop* | S | 1× | 1× | N× | 0.129 | 0.777 | 4.960 | 0.187 | 0.817 | 0.960 | 0.990 |
| Monodepth2-*Boot* | S | N× | N× | 1× | 0.084 | 0.481 | 3.869 | 0.137 | 0.910 | **0.981** | 0.995 |
| Monodepth2-*Snap* | S | 1× | N× | 1× | 0.085 | 0.491 | 3.901 | 0.139 | 0.908 | 0.980 | 0.994 |
| Monodepth2-*Repr* | S | 1× | 1× | 1× | 0.084 | 0.491 | 3.926 | 0.140 | 0.909 | 0.980 | 0.994 |
| Monodepth2-*Log* | S | 1× | 1× | 1× | 0.084 | 0.494 | 3.906 | 0.140 | **0.911** | 0.980 | 0.994 |
| Monodepth2-*Self* | S | (1+1)× | 1× | 1× | **0.083** | 0.475 | 3.854 | 0.137 | **0.911** | 0.980 | 0.994 |
| Monodepth2-*Boot+Log* | S | N× | N× | 1× | 0.084 | 0.483 | 3.873 | 0.137 | 0.909 | **0.981** | 0.995 |
| Monodepth2-*Boot+Self* | S | (1+N)× | N× | 1× | 0.084 | 0.472 | 3.852 | **0.136** | 0.910 | **0.981** | 0.995 |
| Monodepth2-*Snap+Log* | S | 1× | 1× | 1× | 0.084 | 0.488 | 3.894 | 0.138 | **0.911** | **0.981** | 0.994 |
| Monodepth2-*Snap+Self* | S | (1+1)× | 1× | 1× | 0.085 | 0.490 | 3.899 | 0.138 | 0.908 | **0.981** | 0.994 |

Table 5. **Depth evaluation for stereo (S) supervision.** Evaluation on the Eigen test split [2] with improved ground truth [6]. Maximum depth reduced to 50 meters.

| Method | Sup | #Trn | #Par | #Fwd | Abs Rel | Sq Rel | RMSE | RMSE log | $\delta < 1.25$ | $\delta < 1.25^2$ | $\delta < 1.25^3$ |
|---|---|---|---|---|---|---|---|---|---|---|---|
| Monodepth2 [5] | MS | 1× | 1× | 1× | 0.083 | 0.461 | 3.830 | 0.132 | 0.914 | **0.984** | **0.996** |
| Monodepth2-*Post* [5] | MS | 1× | 1× | 2× | **0.082** | **0.445** | 3.790 | **0.130** | 0.915 | **0.984** | **0.996** |
| Monodepth2-*Drop* | MS | 1× | 1× | N× | 0.172 | 1.074 | 5.921 | 0.237 | 0.678 | 0.933 | 0.982 |
| Monodepth2-*Boot* | MS | N× | N× | 1× | 0.086 | 0.485 | 3.925 | 0.137 | 0.906 | 0.981 | 0.995 |
| Monodepth2-*Snap* | MS | 1× | N× | 1× | 0.085 | 0.476 | 3.899 | 0.135 | 0.910 | 0.983 | **0.996** |
| Monodepth2-*Repr* | MS | 1× | 1× | 1× | 0.084 | 0.470 | 3.905 | 0.134 | 0.909 | 0.983 | 0.995 |
| Monodepth2-*Log* | MS | 1× | 1× | 1× | 0.083 | 0.471 | 3.832 | 0.133 | 0.912 | **0.984** | **0.996** |
| Monodepth2-*Self* | MS | (1+1)× | 1× | 1× | 0.083 | 0.455 | **3.781** | **0.130** | 0.915 | **0.984** | **0.996** |
| Monodepth2-*Boot+Log* | MS | N× | N× | 1× | 0.086 | 0.481 | 3.903 | 0.136 | 0.907 | 0.981 | 0.995 |
| Monodepth2-*Boot+Self* | MS | (1+N)× | N× | 1× | 0.085 | 0.462 | 3.815 | 0.132 | 0.911 | 0.983 | **0.996** |
| Monodepth2-*Snap+Log* | MS | 1× | 1× | 1× | 0.084 | 0.481 | 3.900 | 0.134 | 0.911 | 0.982 | **0.996** |
| Monodepth2-*Snap+Self* | MS | (1+1)× | 1× | 1× | 0.084 | 0.467 | 3.810 | 0.132 | 0.912 | 0.983 | **0.996** |

Table 6. **Depth evaluation for monocular+stereo (MS) supervision.** Evaluation on the Eigen test split [2] with improved ground truth [6]. Maximum depth reduced to 50 meters.

## 5. Uncertainty evaluation – 50 meters cap

To complete the experiments from the previous section, we also evaluate uncertainty modelling by assuming a maximum depth of 50 meters. Tables 7, 8 and 9 collects the results of this evaluation for M, S and MS supervisions. Compared with Tables 1, 2 and 3 from the main paper, we highlight that the same behaviour occurs regardless of the maximum depth set to 80 or 50 meters.

| Method | Abs Rel | | RMSE | | $\delta \geq 1.25$ | |
| --- | --- | --- | --- | --- | --- | --- |
| | AUSE | AURG | AUSE | AURG | AUSE | AURG |
| Monodepth2-*Post* | 0.044 | 0.012 | 2.967 | 0.378 | 0.059 | 0.021 |
| Monodepth2-*Drop* | 0.065 | 0.000 | 2.621 | 0.980 | 0.098 | 0.003 |
| Monodepth2-*Boot* | 0.059 | -0.000 | 4.259 | -0.890 | 0.094 | -0.008 |
| Monodepth2-*Snap* | 0.059 | -0.001 | 4.159 | -0.748 | 0.091 | -0.007 |
| Monodepth2-*Repr* | 0.051 | 0.007 | 3.085 | 0.321 | 0.072 | 0.012 |
| Monodepth2-*Log* | 0.038 | 0.020 | 2.547 | 0.908 | 0.047 | 0.039 |
| Monodepth2-*Self* | 0.030 | 0.026 | 2.136 | 1.207 | 0.033 | 0.045 |
| Monodepth2-*Boot+Log* | 0.038 | 0.020 | 2.605 | 0.778 | 0.050 | 0.036 |
| Monodepth2-*Boot+Self* | **0.029** | **0.028** | **2.053** | **1.267** | **0.031** | **0.049** |
| Monodepth2-*Snap+Log* | 0.037 | 0.022 | 2.482 | 0.949 | 0.046 | 0.039 |
| Monodepth2-*Snap+Self* | 0.030 | 0.026 | 2.160 | 1.166 | 0.034 | 0.045 |

Table 7. **Quantitative results for monocular (M) supervision.** Evaluation on the Eigen test split [2] with improved ground truth [6]. Maximum depth cap to 50 meters.

| Method | Abs Rel | | RMSE | | $\delta \geq 1.25$ | |
| --- | --- | --- | --- | --- | --- | --- |
| | AUSE | AURG | AUSE | AURG | AUSE | AURG |
| Monodepth2-*Post* | 0.036 | 0.019 | 2.662 | 0.663 | 0.048 | 0.033 |
| Monodepth2-*Drop* | 0.102 | -0.029 | 6.276 | -2.234 | 0.240 | -0.086 |
| Monodepth2-*Boot* | 0.028 | 0.029 | 2.555 | 0.790 | 0.037 | 0.046 |
| Monodepth2-*Snap* | 0.028 | 0.029 | 2.450 | 0.925 | 0.035 | 0.049 |
| Monodepth2-*Repr* | 0.040 | 0.016 | 2.386 | 1.011 | 0.052 | 0.031 |
| Monodepth2-*Log* | 0.022 | 0.035 | 0.980 | 2.401 | 0.019 | 0.063 |
| Monodepth2-*Self* | 0.022 | 0.034 | 1.858 | 1.479 | 0.026 | 0.055 |
| Monodepth2-*Boot+Log* | **0.020** | **0.037** | **0.847** | **2.504** | **0.018** | **0.066** |
| Monodepth2-*Boot+Self* | 0.023 | 0.034 | 1.795 | 1.533 | 0.025 | 0.058 |
| Monodepth2-*Snap+Log* | 0.021 | 0.036 | 0.929 | 2.443 | 0.019 | 0.064 |
| Monodepth2-*Snap+Self* | 0.023 | 0.034 | 1.863 | 1.506 | 0.027 | 0.057 |

Table 8. **Quantitative results for stereo (S) supervision.** Evaluation on the Eigen test split [2] with improved ground truth [6]. Maximum depth cap to 50 meters.

| Method | Abs Rel | | RMSE | | $\delta \geq 1.25$ | |
| --- | --- | --- | --- | --- | --- | --- |
| | AUSE | AURG | AUSE | AURG | AUSE | AURG |
| Monodepth2-*Post* | 0.036 | 0.018 | 2.650 | 0.621 | 0.048 | 0.031 |
| Monodepth2-*Drop* | 0.103 | -0.027 | 7.188 | -2.621 | 0.307 | -0.085 |
| Monodepth2-*Boot* | 0.029 | 0.029 | 2.470 | 0.914 | 0.038 | 0.049 |
| Monodepth2-*Snap* | 0.028 | 0.028 | 2.413 | 0.947 | 0.037 | 0.047 |
| Monodepth2-*Repr* | 0.046 | 0.010 | 2.781 | 0.587 | 0.065 | 0.018 |
| Monodepth2-*Log* | 0.028 | 0.028 | **1.818** | **1.495** | 0.030 | 0.051 |
| Monodepth2-*Self* | **0.023** | **0.033** | 1.870 | 1.391 | **0.027** | 0.052 |
| Monodepth2-*Boot+Log* | 0.030 | 0.028 | 2.112 | 1.256 | 0.035 | 0.052 |
| Monodepth2-*Boot+Self* | **0.023** | **0.033** | 1.880 | 1.407 | **0.027** | **0.056** |
| Monodepth2-*Snap+Log* | 0.030 | 0.026 | 2.152 | 1.218 | 0.035 | 0.048 |
| Monodepth2-*Snap+Self* | **0.023** | **0.033** | 1.880 | 1.403 | **0.027** | 0.054 |

Table 9. **Quantitative results for monocular+stereo (MS) supervision.** Evaluation on the Eigen test split [2] with improved ground truth [6]. Maximum depth cap to 50 meters.

# 6. Sparsification curves – 50 meters cap

To conclude the evaluation at 50 meters, we report sparsification curves. Figures 4, 5 and 6 confirm that for M, S and MS the same behavior observed for the 80 meters evaluation is kept.

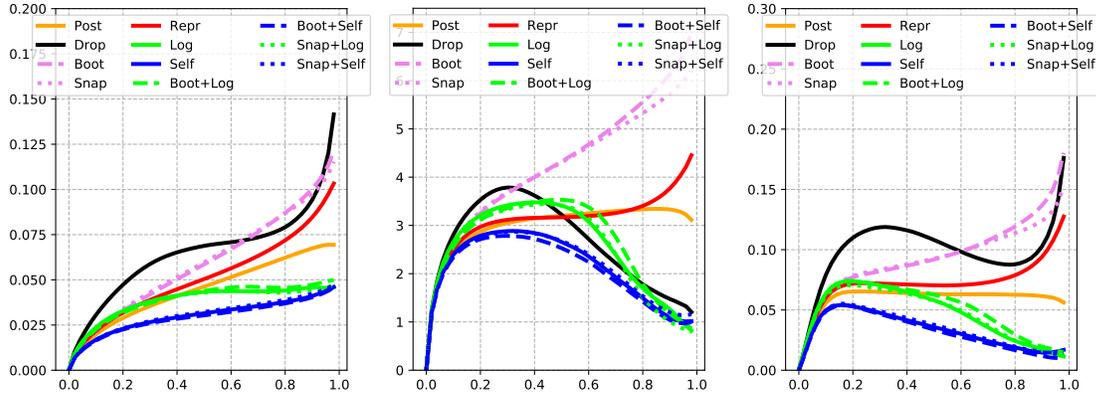

Figure 4. **Sparsification Error curves for monocular (M) supervision.** From left to right, Abs Rel, RMSE and $\delta \geq 1.25$.

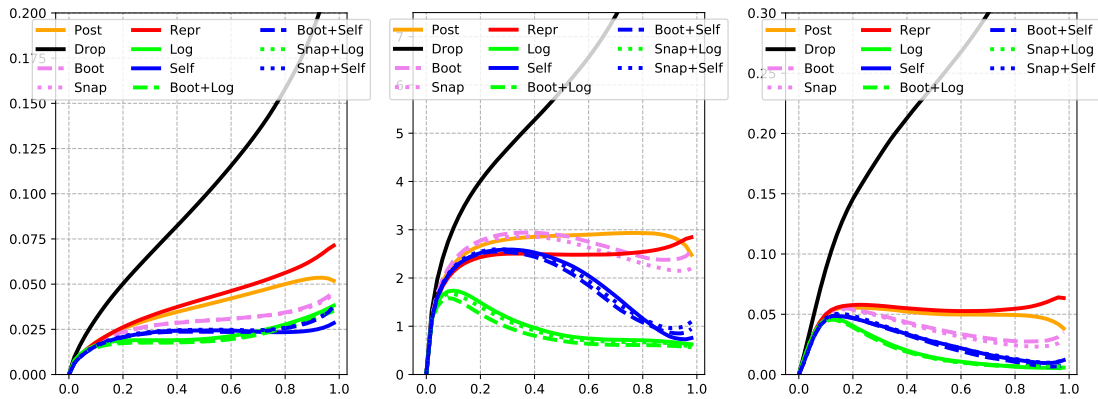

Figure 5. **Sparsification Error curves for stereo (S) supervision.** From left to right, Abs Rel, RMSE and $\delta \geq 1.25$.

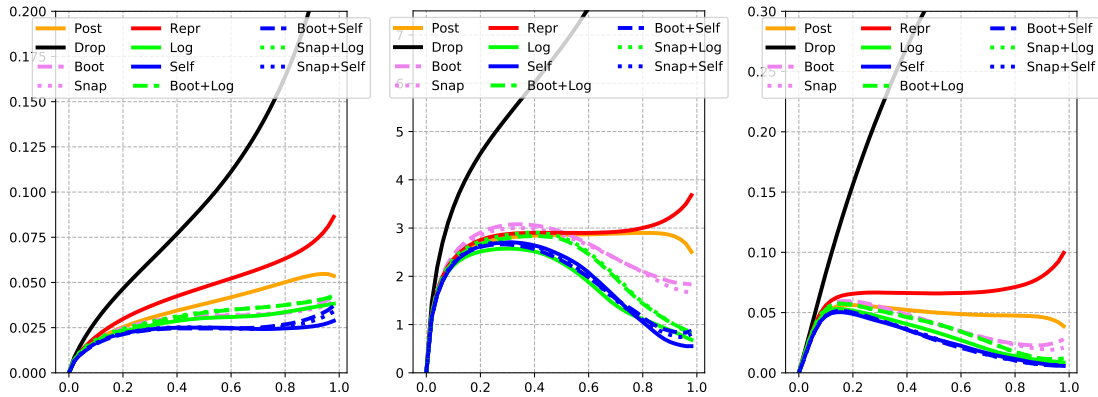

Figure 6. **Sparsification Error curves for monocular+stereo (MS) supervision.** From left to right, Abs Rel, RMSE and $\delta \geq 1.25$.

# 7. Depth evaluation – raw LiDAR (80 meters)

To ease comparison with previous works in literature [5], we also report the same evaluation carried out in the main paper by assuming the raw LiDAR depth measurements as ground truth. Tables 10, 11 and 12 collect the outcome of this evaluation.

| Method | Sup | #Trn | #Par | #Fwd | Abs Rel | Sq Rel | RMSE | RMSE log | $\delta < 1.25$ | $\delta < 1.25^2$ | $\delta < 1.25^3$ |
|---|---|---|---|---|---|---|---|---|---|---|---|
| Monodepth2 [5]        | M | 1×      | 1× | 1× | 0.115 | 0.903 | 4.863 | 0.193 | 0.877 | 0.959 | 0.981 |
| Monodepth2-*Post* [5] | M | 1×      | 1× | 2× | 0.112 | 0.852 | 4.755 | 0.190 | **0.881** | 0.960 | 0.981 |
| Monodepth2-*Drop*     | M | 1×      | 1× | N× | 0.126 | 0.895 | 4.911 | 0.198 | 0.850 | 0.952 | 0.982 |
| Monodepth2-*Boot*     | M | N×      | N× | 1× | 0.114 | 0.803 | **4.639** | 0.187 | 0.875 | **0.961** | **0.983** |
| Monodepth2-*Snap*     | M | 1×      | N× | 1× | 0.114 | 0.865 | 4.787 | 0.190 | 0.877 | 0.960 | 0.982 |
| Monodepth2-*Repr*     | M | 1×      | 1× | 1× | 0.116 | 0.914 | 4.853 | 0.193 | 0.875 | 0.958 | 0.981 |
| Monodepth2-*Log*      | M | 1×      | 1× | 1× | 0.113 | 0.928 | 4.919 | 0.192 | 0.876 | 0.958 | 0.981 |
| Monodepth2-*Self*     | M | (1+1)×  | 1× | 1× | **0.111** | 0.863 | 4.756 | 0.188 | **0.881** | **0.961** | 0.982 |
| Monodepth2-*Boot+Log* | M | N×      | N× | 1× | 0.114 | **0.797** | 4.661 | 0.186 | 0.872 | 0.960 | **0.983** |
| Monodepth2-*Boot+Self*| M | (1+N)×  | N× | 1× | **0.111** | 0.826 | 4.667 | **0.184** | 0.880 | **0.961** | **0.983** |
| Monodepth2-*Snap+Log* | M | 1×      | 1× | 1× | 0.117 | 0.900 | 4.838 | 0.192 | 0.873 | 0.958 | 0.981 |
| Monodepth2-*Snap+Self*| M | (1+1)×  | 1× | 1× | 0.112 | 0.871 | 4.747 | 0.187 | 0.880 | **0.961** | 0.982 |

Table 10. **Depth evaluation for monocular (M) supervision.** Evaluation on the Eigen test split [2] with raw LiDAR.

| Method | Sup | #Trn | #Par | #Fwd | Abs Rel | Sq Rel | RMSE | RMSE log | $\delta < 1.25$ | $\delta < 1.25^2$ | $\delta < 1.25^3$ |
|---|---|---|---|---|---|---|---|---|---|---|---|
| Monodepth2 [5]        | S | 1×     | 1× | 1× | 0.109 | 0.873 | 4.960 | 0.209 | 0.864 | 0.948 | 0.975 |
| Monodepth2-*Post* [5] | S | 1×     | 1× | 2× | 0.108 | 0.842 | 4.892 | 0.207 | 0.866 | 0.949 | 0.976 |
| Monodepth2-*Drop*     | S | 1×     | 1× | N× | 0.151 | 1.110 | 5.780 | 0.244 | 0.764 | 0.926 | 0.970 |
| Monodepth2-*Boot*     | S | N×     | N× | 1× | 0.108 | 0.822 | 4.809 | 0.201 | 0.866 | 0.951 | 0.977 |
| Monodepth2-*Snap*     | S | 1×     | N× | 1× | 0.109 | 0.868 | 4.918 | 0.206 | 0.864 | 0.949 | 0.976 |
| Monodepth2-*Repr*     | S | 1×     | 1× | 1× | 0.109 | 0.876 | 4.975 | 0.210 | 0.862 | 0.948 | 0.975 |
| Monodepth2-*Log*      | S | 1×     | 1× | 1× | 0.110 | 0.876 | 4.952 | 0.209 | 0.865 | 0.948 | 0.975 |
| Monodepth2-*Self*     | S | (1+1)× | 1× | 1× | 0.109 | 0.858 | 4.920 | 0.206 | **0.867** | 0.949 | 0.976 |
| Monodepth2-*Boot+Log* | S | N×     | N× | 1× | **0.107** | 0.811 | **4.796** | 0.200 | 0.866 | **0.952** | **0.978** |
| Monodepth2-*Boot+Self*| S | (1+N)× | N× | 1× | **0.107** | 0.806 | 4.798 | **0.199** | 0.866 | **0.952** | **0.978** |
| Monodepth2-*Snap+Log* | S | 1×     | 1× | 1× | 0.108 | 0.851 | 4.894 | 0.204 | **0.867** | 0.951 | 0.976 |
| Monodepth2-*Snap+Self*| S | (1+1)× | 1× | 1× | 0.109 | 0.848 | 4.895 | 0.204 | 0.864 | 0.950 | 0.977 |

Table 11. **Depth evaluation for stereo (S) supervision.** Evaluation on the Eigen test split [2] with raw LiDAR.

| Method | Sup | #Trn | #Par | #Fwd | Abs Rel | Sq Rel | RMSE | RMSE log | $\delta < 1.25$ | $\delta < 1.25^2$ | $\delta < 1.25^3$ |
|---|---|---|---|---|---|---|---|---|---|---|---|
| Monodepth2 [5]        | MS | 1×     | 1× | 1× | 0.106 | 0.818 | 4.750 | 0.196 | 0.874 | 0.957 | 0.979 |
| Monodepth2-*Post* [5] | MS | 1×     | 1× | 2× | **0.104** | 0.787 | 4.687 | 0.194 | **0.876** | 0.958 | 0.980 |
| Monodepth2-*Drop*     | MS | 1×     | 1× | N× | 0.201 | 1.421 | 6.704 | 0.295 | 0.593 | 0.896 | 0.962 |
| Monodepth2-*Boot*     | MS | N×     | N× | 1× | 0.109 | 0.787 | 4.747 | 0.195 | 0.866 | 0.956 | 0.980 |
| Monodepth2-*Snap*     | MS | 1×     | N× | 1× | 0.109 | 0.828 | 4.815 | 0.198 | 0.869 | 0.956 | 0.979 |
| Monodepth2-*Repr*     | MS | 1×     | 1× | 1× | 0.109 | 0.820 | 4.830 | 0.199 | 0.869 | 0.955 | 0.979 |
| Monodepth2-*Log*      | MS | 1×     | 1× | 1× | 0.107 | 0.839 | 4.792 | 0.197 | 0.873 | 0.956 | 0.979 |
| Monodepth2-*Self*     | MS | (1+1)× | 1× | 1× | **0.104** | 0.797 | 4.686 | 0.192 | **0.876** | 0.957 | 0.980 |
| Monodepth2-*Boot+Log* | MS | N×     | N× | 1× | 0.108 | 0.784 | 4.735 | 0.194 | 0.866 | 0.955 | 0.980 |
| Monodepth2-*Boot+Self*| MS | (1+N)× | N× | 1× | 0.105 | **0.766** | **4.638** | **0.189** | 0.873 | **0.958** | **0.982** |
| Monodepth2-*Snap+Log* | MS | 1×     | 1× | 1× | 0.108 | 0.824 | 4.821 | 0.196 | 0.870 | 0.956 | 0.980 |
| Monodepth2-*Snap+Self*| MS | (1+1)× | 1× | 1× | 0.105 | 0.795 | 4.682 | 0.191 | 0.875 | 0.957 | 0.981 |

Table 12. **Depth evaluation for monocular+stereo (MS) supervision.** Evaluation on the Eigen test split [2] with raw LiDAR.

# 8. Uncertainty evaluation – raw LiDAR (80 meters)

We report the evaluation of uncertainty modelling adopting the raw LiDAR as well. Tables 13, 14 and 15 resume the outcome showing how the same behaviour occurs, *i.e. Self* solutions are much better when dealing with M and MS supervisions, while *Log* outperforms *Self* approach training with S.

|  | Abs Rel | | RMSE | | $\delta \geq 1.25$ | |
| --- | --- | --- | --- | --- | --- | --- |
| Method | AUSE | AURG | AUSE | AURG | AUSE | AURG |
| Monodepth2-*Post* | 0.053 | 0.020 | 3.322 | 0.734 | 0.069 | 0.040 |
| Monodepth2-*Drop* | 0.083 | 0.001 | 3.044 | 1.142 | 0.133 | 0.001 |
| Monodepth2-*Boot* | 0.064 | 0.010 | 4.266 | -0.334 | 0.091 | 0.023 |
| Monodepth2-*Snap* | 0.068 | 0.007 | 4.391 | -0.315 | 0.095 | 0.018 |
| Monodepth2-*Repr* | 0.058 | 0.019 | 3.282 | 0.855 | 0.080 | 0.034 |
| Monodepth2-*Log* | 0.051 | 0.027 | 3.097 | 1.188 | 0.060 | 0.056 |
| Monodepth2-*Self* | 0.036 | 0.038 | 2.292 | 1.779 | 0.037 | 0.072 |
| Monodepth2-*Boot+Log* | 0.046 | 0.028 | 2.830 | 1.119 | 0.060 | 0.057 |
| Monodepth2-*Boot+Self* | **0.033** | **0.040** | **2.124** | **1.857** | **0.033** | **0.077** |
| Monodepth2-*Snap+Log* | 0.047 | 0.030 | 2.837 | 1.281 | 0.058 | 0.059 |
| Monodepth2-*Snap+Self* | 0.036 | 0.038 | 2.331 | 1.725 | 0.036 | 0.073 |

Table 13. **Quantitative results for monocular (M) supervision.** Evaluation on the Eigen test split [2] with raw LiDAR.

|  | Abs Rel | | RMSE | | $\delta \geq 1.25$ | |
| --- | --- | --- | --- | --- | --- | --- |
| Method | AUSE | AURG | AUSE | AURG | AUSE | AURG |
| Monodepth2-*Post* | 0.044 | 0.029 | 3.104 | 1.119 | 0.063 | 0.057 |
| Monodepth2-*Drop* | 0.114 | -0.027 | 6.821 | -2.088 | 0.266 | -0.071 |
| Monodepth2-*Boot* | 0.032 | 0.041 | 2.640 | 1.504 | 0.041 | 0.080 |
| Monodepth2-*Snap* | 0.034 | 0.041 | 2.654 | 1.589 | 0.042 | 0.080 |
| Monodepth2-*Repr* | 0.045 | 0.029 | 2.647 | 1.650 | 0.069 | 0.055 |
| Monodepth2-*Log* | 0.026 | 0.048 | 1.144 | 3.126 | 0.031 | 0.091 |
| Monodepth2-*Self* | 0.026 | 0.048 | 1.931 | 2.321 | 0.029 | 0.091 |
| Monodepth2-*Boot+Log* | **0.024** | **0.049** | **0.988** | **3.151** | 0.028 | **0.093** |
| Monodepth2-*Boot+Self* | 0.025 | 0.048 | 1.813 | 2.326 | **0.027** | **0.093** |
| Monodepth2-*Snap+Log* | 0.025 | **0.049** | 1.082 | 3.148 | 0.029 | 0.092 |
| Monodepth2-*Snap+Self* | 0.026 | 0.048 | 1.937 | 2.286 | 0.030 | 0.092 |

Table 14. **Quantitative results for stereo (S) supervision.** Evaluation on the Eigen test split [2] with raw LiDAR.

|  | Abs Rel | | RMSE | | $\delta \geq 1.25$ | |
| --- | --- | --- | --- | --- | --- | --- |
| Method | AUSE | AURG | AUSE | AURG | AUSE | AURG |
| Monodepth2-*Post* | 0.043 | 0.028 | 3.021 | 1.024 | 0.060 | 0.053 |
| Monodepth2-*Drop* | 0.110 | -0.022 | 7.450 | -2.370 | 0.321 | -0.065 |
| Monodepth2-*Boot* | 0.032 | 0.040 | 2.590 | 1.477 | 0.043 | 0.078 |
| Monodepth2-*Snap* | 0.034 | 0.039 | 2.654 | 1.482 | 0.044 | 0.075 |
| Monodepth2-*Repr* | 0.051 | 0.021 | 3.026 | 1.120 | 0.078 | 0.039 |
| Monodepth2-*Log* | 0.032 | 0.040 | 2.093 | 2.039 | 0.040 | 0.075 |
| Monodepth2-*Self* | **0.025** | 0.046 | 1.896 | **2.153** | 0.029 | 0.083 |
| Monodepth2-*Boot+Log* | 0.034 | 0.039 | 2.331 | 1.735 | 0.045 | 0.076 |
| Monodepth2-*Boot+Self* | **0.025** | **0.047** | **1.859** | 2.131 | **0.028** | **0.087** |
| Monodepth2-*Snap+Log* | 0.035 | 0.037 | 2.431 | 1.719 | 0.046 | 0.072 |
| Monodepth2-*Snap+Self* | **0.025** | **0.047** | 1.903 | 2.137 | 0.029 | 0.085 |

Table 15. **Quantitative results for monocular+stereo (MS) supervision.** Evaluation on the Eigen test split [2] with raw LiDAR.

# 9. Sparsification curves – raw LiDAR (80 meters)

We also report Sparsification Error curves to perceive the behaviour of the modelled uncertainties better. Figures 7, 8 and 9 highlight once more how the variants based on *Self* outperforms *Log* ones on M and MS, with these latter yielding better results on S in particular when considering RMSE sparsification.

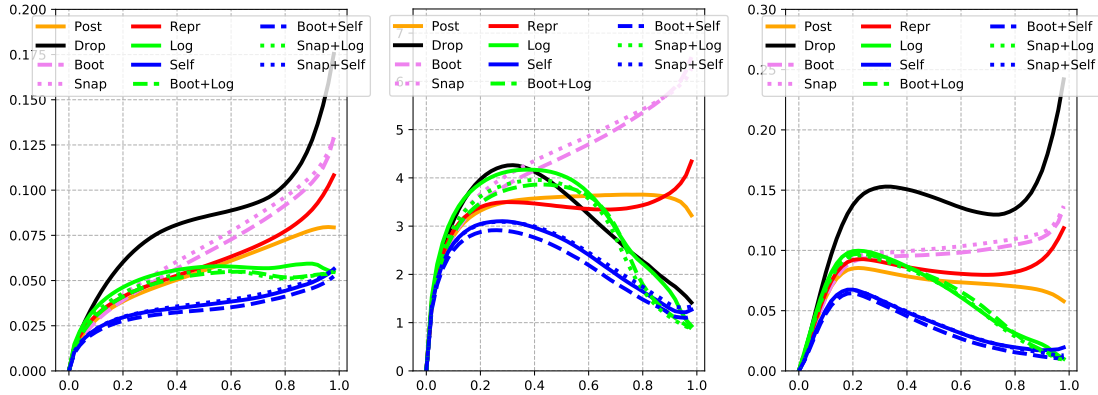

Figure 7. **Sparsification Error curves for monocular (M) supervision.** From left to right, Abs Rel, RMSE and $\delta \geq 1.25$.

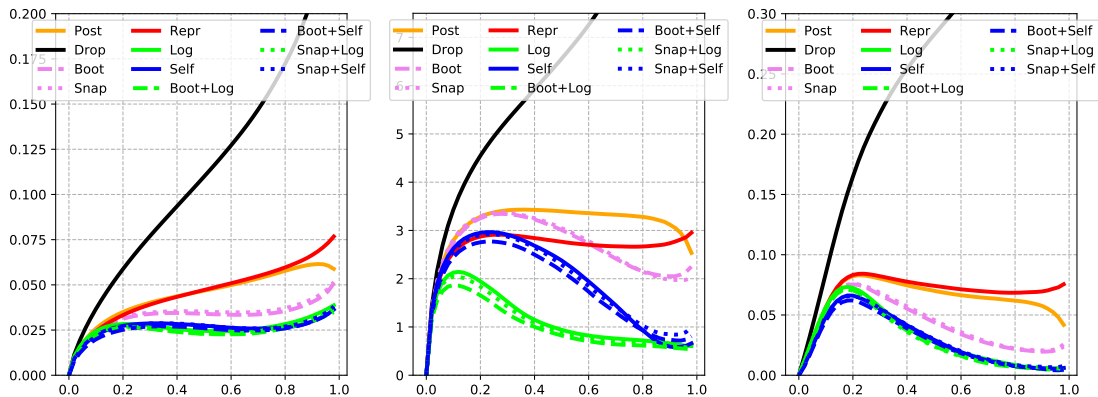

Figure 8. **Sparsification Error curves for stereo (S) supervision.** From left to right, Abs Rel, RMSE and $\delta \geq 1.25$.

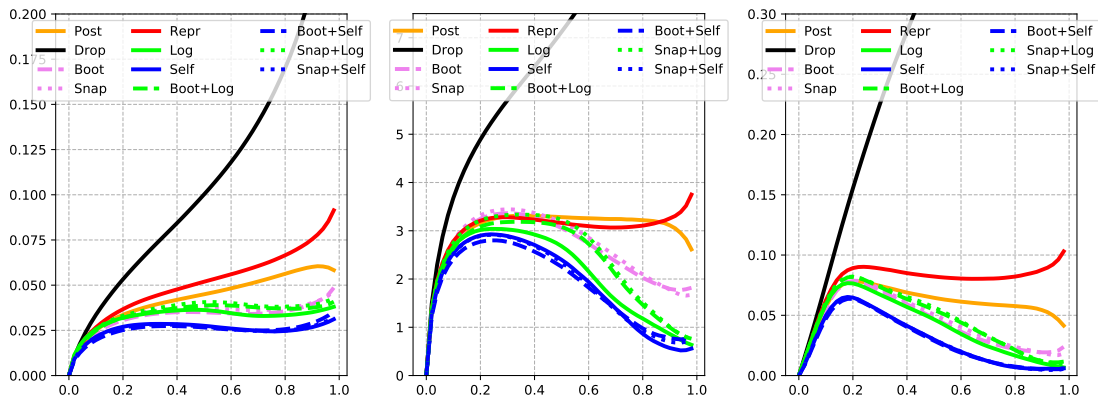

Figure 9. **Sparsification Error curves for monocular+stereo (MS) supervision.** From left to right, Abs Rel, RMSE and $\delta \geq 1.25$.

9

## 10. Qualitative results

Finally, we report some qualitative examples of both depth and uncertainty maps obtained by the different methods evaluated in the paper. Given the high amount of images produced by all the considered variants, at first, we introduce the notation to ease readability.

### 10.1. Colormap encodings

To show qualitative examples obtained by our framework, we adopt colormap **magma** for depth maps and colormap **hot** for uncertainty. Figure 10 shows the adopted colormaps and how they range from far to close depth and from low to high uncertainty.

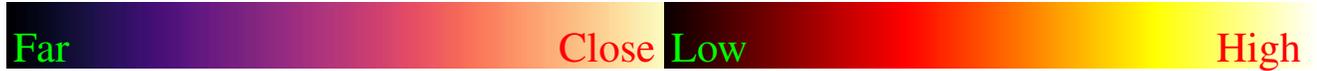

Figure 10. **Colormap encodings for depth and uncertainty.** We choose colormap **magma** (on left) to encode depth maps and colormap **hot** (on right) for uncertainty. Best viewed with colors.

### 10.2. Results topology

In our paper, we evaluated eleven different strategies to obtain depth and corresponding uncertainty maps. Thus, we report both outcomes for each of the considered variants, organized as shown in Figure 11.

| Monodepth2 | Reference Image | Post | Post |
|---|---|---|---|
| Drop | Drop | Boot | Boot |
| Snap | Snap | Repr | Repr |
| Log | Log | Self | Self |
| Boot+Log | Boot+Log | Boot+Self | Boot+Self |
| Snap+Log | Snap+Log | Snap+Self | Snap+Self |

Figure 11. **Legend for qualitative results.** Each cell in the table shows what each of the qualitative figure reported in the reminder represents.

We will show results on three images taken from the Eigen test split [2], respectively **2011_09_26_drive_0002_sync/0000000021**, **2011_09_26_drive_0013_sync/0000000045** and **2011_09_26_drive_0101_sync/0000000114**. For each one, we report results for the network trained with monocular (M), stereo (S) or both (MS) supervision strategies.

## 10.3. Image 2011_09_26_drive_0002_sync/0000000021

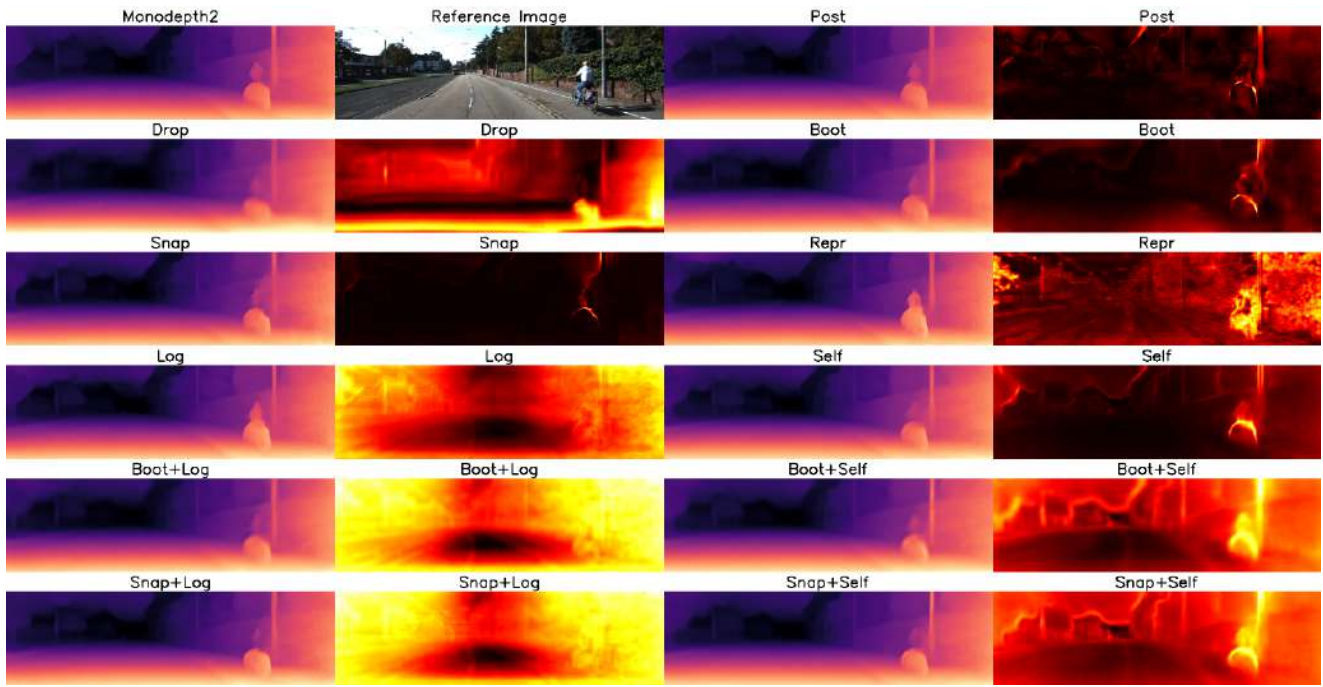

Figure 12. **Qualitative results on image 2011_09_26_drive_0002_sync/0000000021 from the Eigen test split [2].** Each row shows depth and uncertainty maps from one of the considered variants trained with monocular (M) supervision.

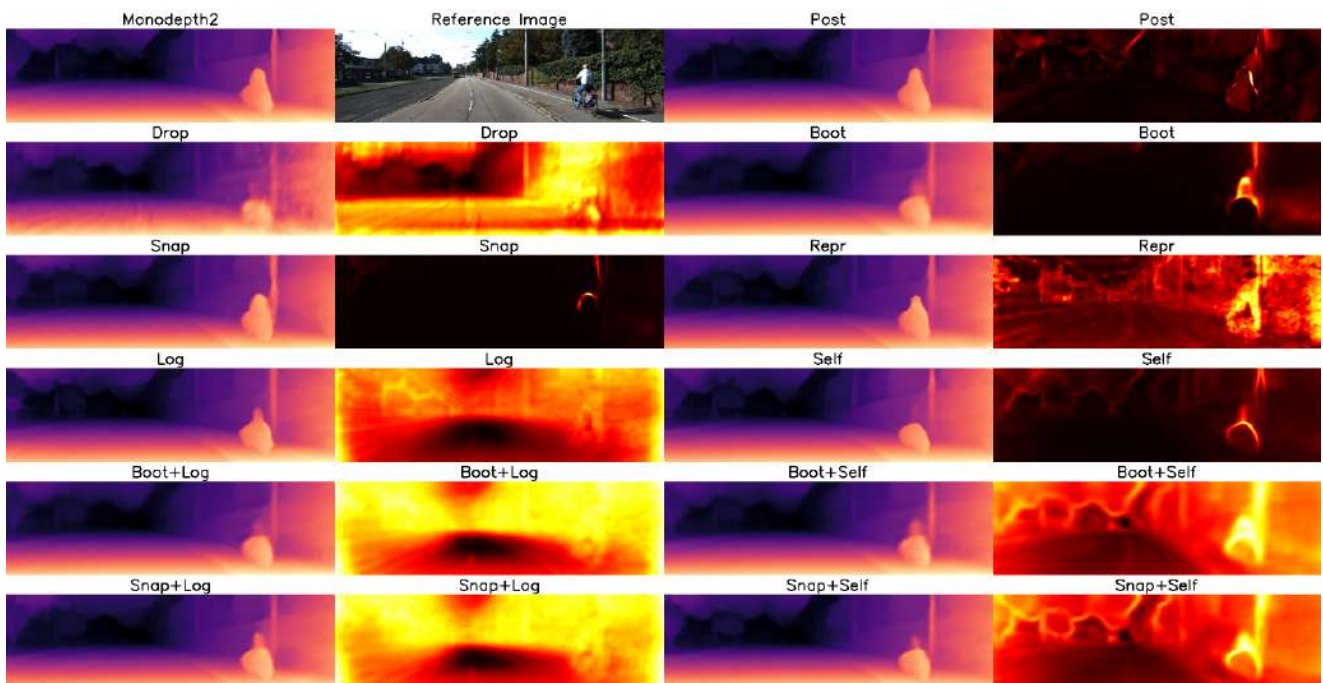

Figure 13. **Qualitative results on image 2011_09_26_drive_0002_sync/0000000021 from the Eigen test split [2].** Each row shows depth and uncertainty maps from one of the considered variants trained with stereo (S) supervision.

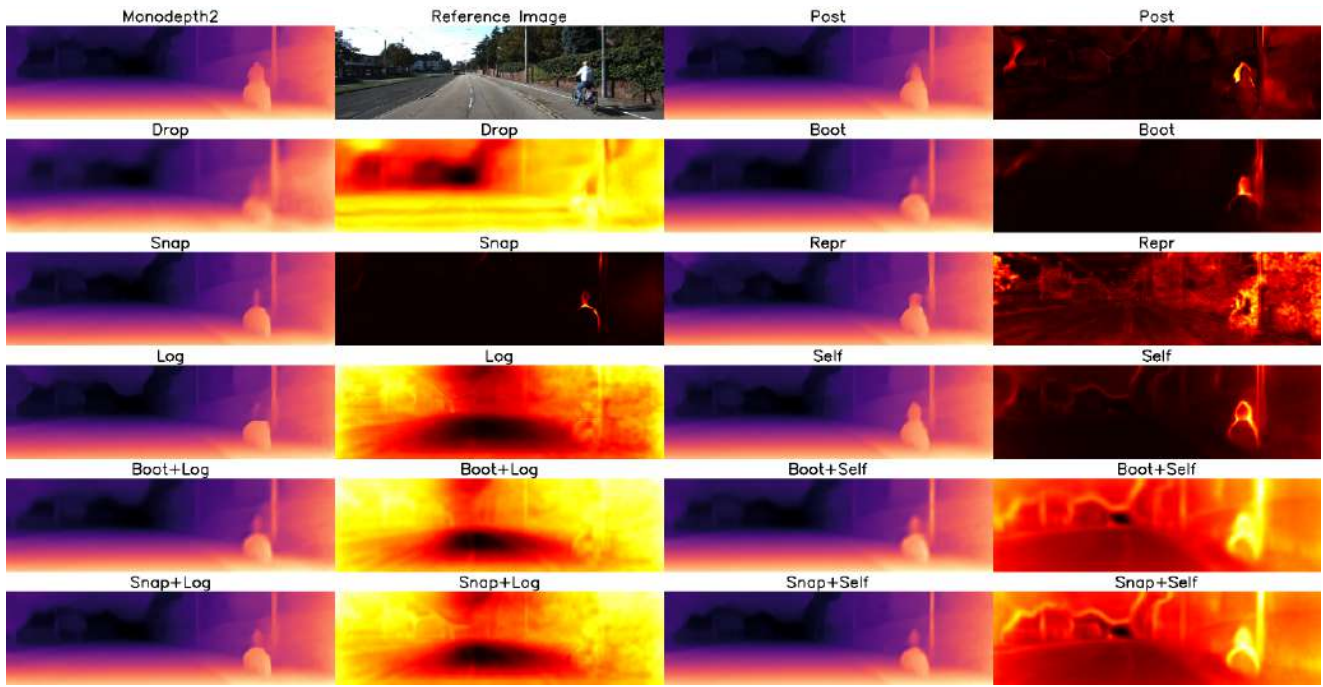

Figure 14. **Qualitative results on image 2011_09_26_drive_0002_sync/0000000021 from the Eigen test split [2].** Each row shows depth and uncertainty maps from one of the considered variants trained with mono+stereo (MS) supervision.

## 10.4. Image 2011_09_26_drive_0013_sync/0000000045

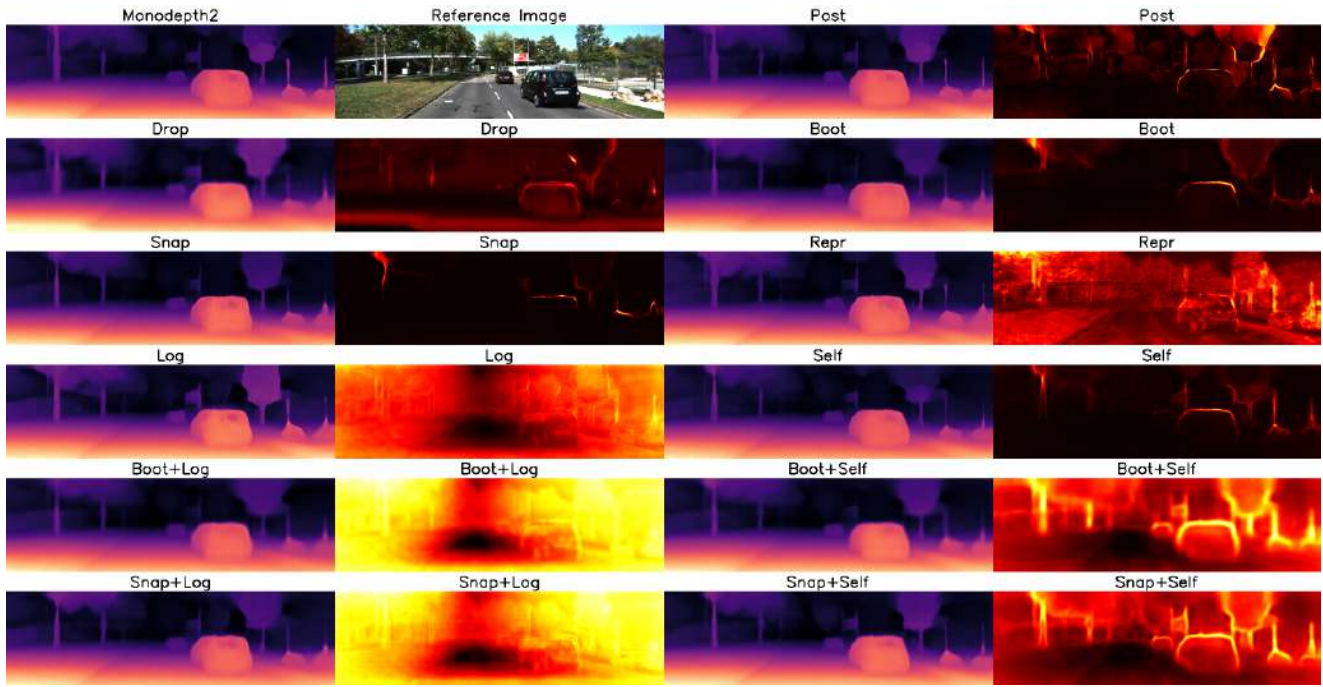

Figure 15. **Qualitative results on image 2011_09_26_drive_0013_sync/0000000045 from the Eigen test split [2].** Each row shows depth and uncertainty maps from one of the considered variants trained with monocular (M) supervision.

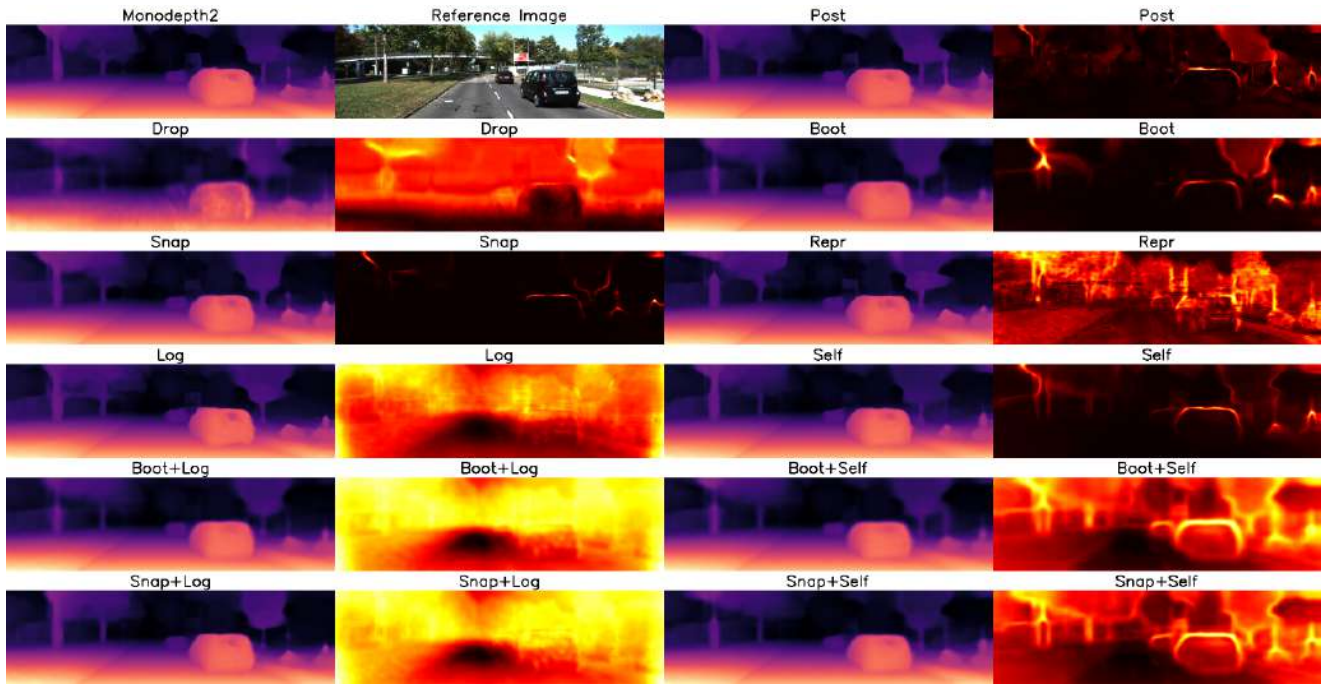

Figure 16. **Qualitative results on image 2011_09_26_drive_0013_sync/0000000045 from the Eigen test split [2].** Each row shows depth and uncertainty maps from one of the considered variants trained with stereo (S) supervision.

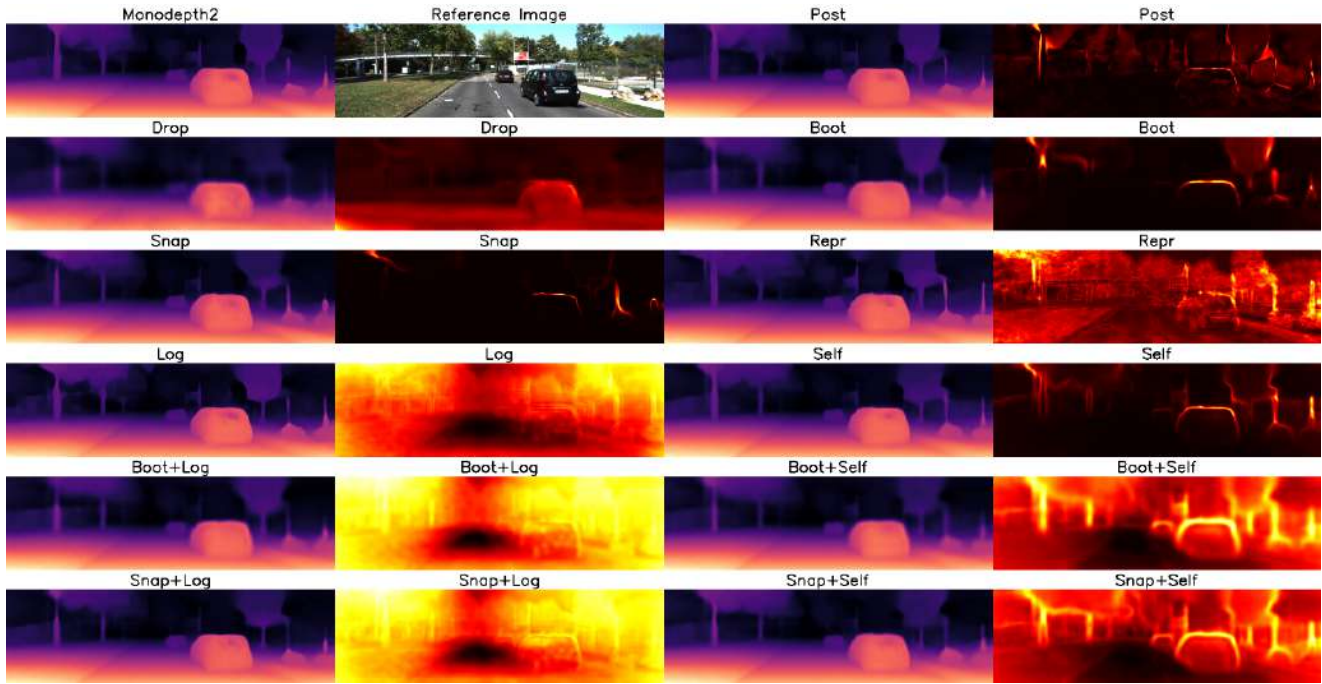

Figure 17. **Qualitative results on image 2011_09_26_drive_0013_sync/0000000045 from the Eigen test split [2].** Each row shows depth and uncertainty maps from one of the considered variants trained with mono+stereo (MS) supervision.

13

## 10.5. Image 2011_09_26_drive_0101_sync/0000000114

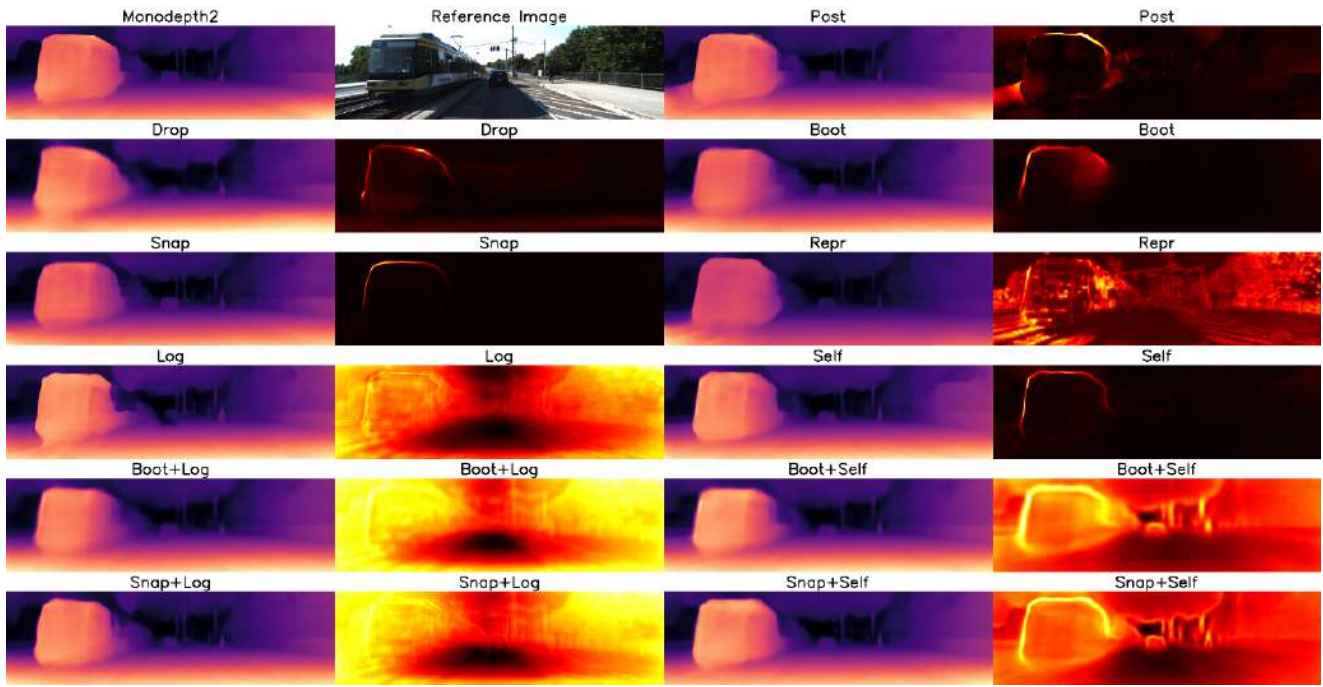

Figure 18. **Qualitative results on image 2011_09_26_drive_0101_sync/0000000114 from the Eigen test split [2].** Each row shows depth and uncertainty maps from one of the considered variants trained with monocular (M) supervision.

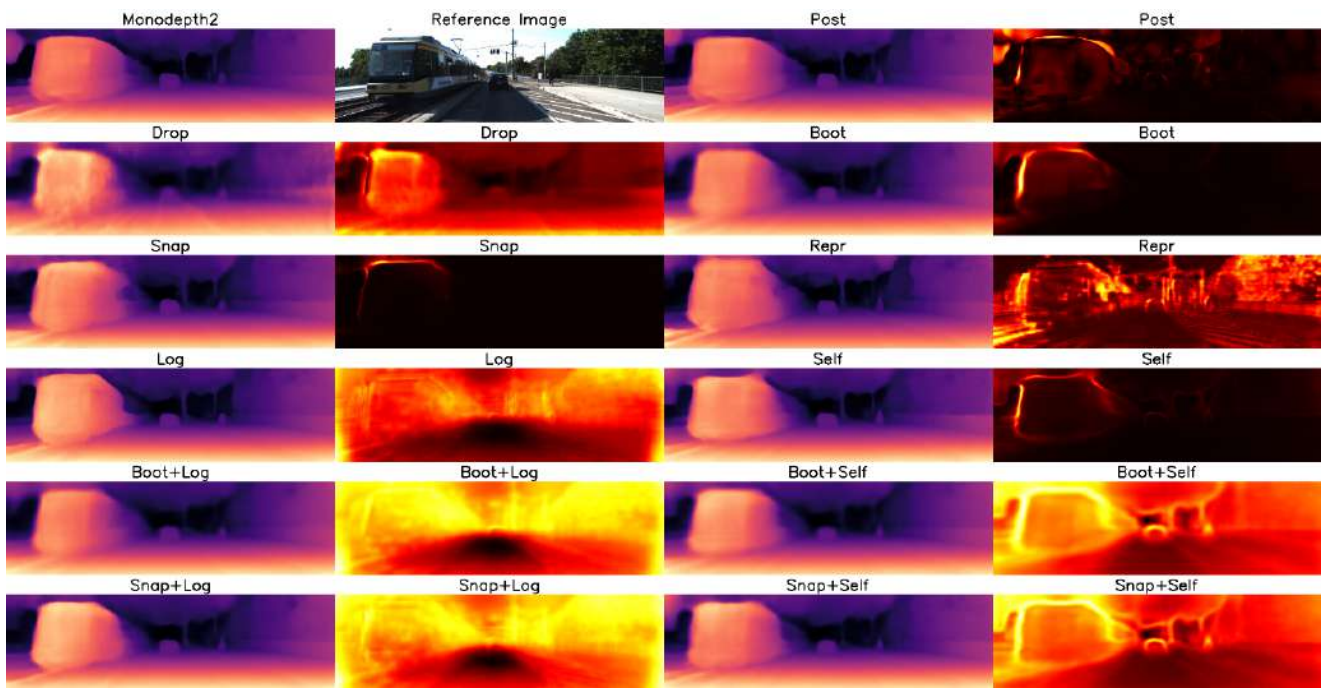

Figure 19. **Qualitative results on image 2011_09_26_drive_0101_sync/0000000114 from the Eigen test split [2].** Each row shows depth and uncertainty maps from one of the considered variants trained with stereo (S) supervision.

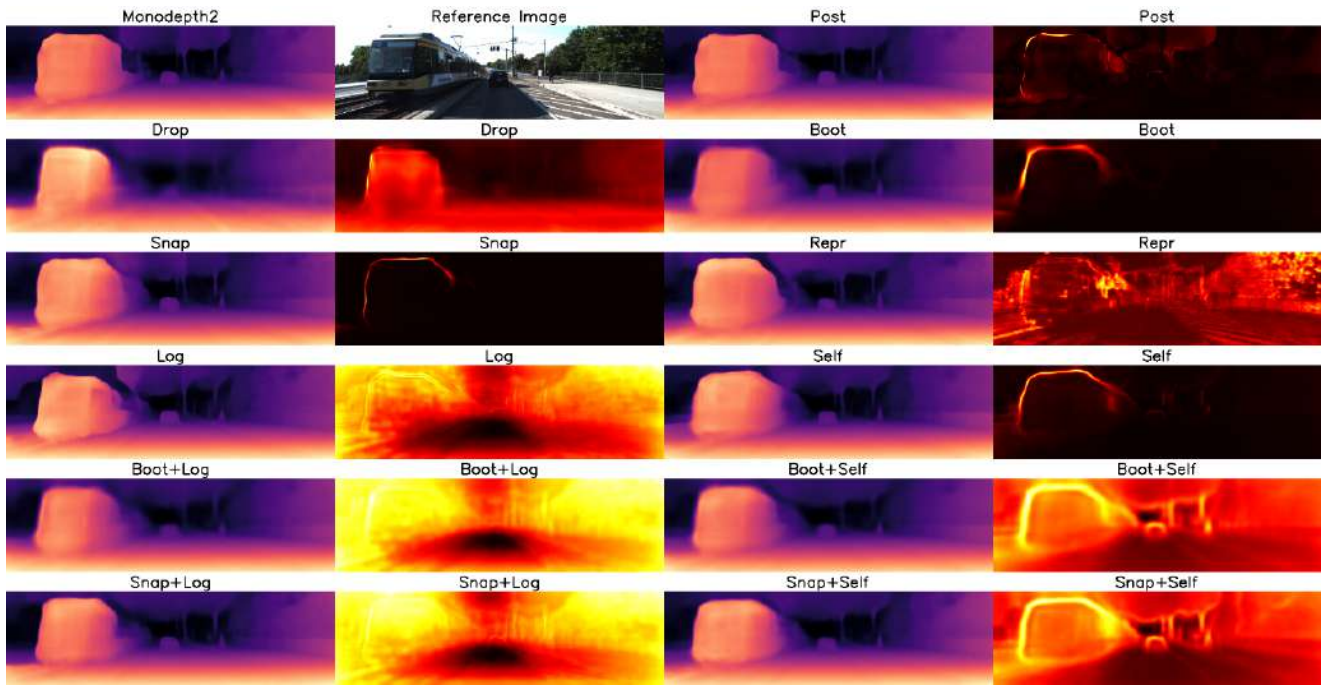

Figure 20. **Qualitative results on image 2011_09_26_drive_0101_sync/0000000114 from the Eigen test split [2].** Each row shows depth and uncertainty maps from one of the considered variants trained with mono+stereo (MS) supervision.

## 10.6. Qualitative video sequence

Finally, we refer the reader to the supplementary video available at www.youtube.com/watch?v=bxVPXqf4zt4, featuring the **2011_09_26_drive_0101_sync** sequence from the KITTI dataset and showing in order results for M, S and MS supervisions. From the video, we can perceive some of the behaviours highlighted in the submitted paper and this document. Specifically, we can observe how *Drop* provides reasonable uncertainty estimation when trained with M while it fails with S and MS. Moreover, we can notice how *Log* estimates are much more defined when dealing with S supervision compared to M and MS. Finally, the video also highlights how *Self* solutions are much more selective at providing high uncertainties compared to *Log* ones.


}

\end{document}